\documentclass[10pt,twocolumn,letterpaper]{article}

\usepackage[applications]{wacv}              

\usepackage{siunitx}
\usepackage{wrapfig}
\usepackage[accsupp]{axessibility}

\definecolor{wacvblue}{rgb}{0.21,0.49,0.74}
\usepackage[pagebackref,breaklinks,colorlinks,allcolors=wacvblue]{hyperref}

\newcommand{\phpf}{\textit{PHerc.~Paris~4}}
\newcommand{\phost}{\textit{PHerc.~172}}

\newcommand{\notarxiv}[1]{}

\renewcommand{\paragraph}[1]{\vspace{0.5pt}\par\noindent\textbf{#1}~}

\def\wacvPaperID{1426}
\def\confName{WACV}
\def\confYear{2026}

\title{Virtually Unrolling the Herculaneum Papyri by Diffeomorphic Spiral Fitting}

\author{Paul Henderson\\
University of Glasgow\\
Scotland, UK\\
{\tt\small paul@pmh47.net}%
}

\begin{document}

\setlength\floatsep{6pt}
\setlength\textfloatsep{6pt}

\setlength\abovedisplayskip{2pt}
\setlength\belowdisplayskip{2pt}
\setlength\abovedisplayshortskip{0pt}
\setlength\belowdisplayshortskip{1pt}

\maketitle

\begin{abstract}
The Herculaneum Papyri are a collection of rolled papyrus documents that were charred and buried by the famous eruption of Mount Vesuvius.
They promise to contain a wealth of previously unseen Greek and Latin texts, but are extremely fragile and thus most cannot be unrolled physically.
A solution to access these texts is virtual unrolling, where the papyrus surface is digitally traced out in a CT scan of the scroll, to create a flattened representation.
This tracing is very laborious to do manually in gigavoxel-sized scans, so automated approaches are desirable.
We present the first top-down method that automatically fits a surface model to a CT scan of a severely damaged scroll.
We take a novel approach that globally fits an explicit parametric model of the deformed scroll to existing neural network predictions of where the rolled papyrus likely passes.
Our method guarantees the resulting surface is a single continuous 2D sheet, even passing through regions where the surface is not detectable in the CT scan.
We conduct comprehensive experiments on high-resolution CT scans of two scrolls, showing that our approach successfully unrolls large regions, and exceeds the performance of the only existing automated unrolling method suitable for this data.
\end{abstract}

\section{Introduction}
\label{sec:introduction}

Buried for almost 2000 years, the Herculaneum Papyri are a trove of classical texts, many of which remain unread \cite{sider05book}.
They originally formed a small private library, but during the famous eruption of Mount Vesuvius in \textsc{ad}79, the tightly rolled papyrus scrolls were carbonized by the high temperatures, then buried and badly distorted over the centuries.

Despite being rediscovered and excavated starting in 1752, many of the scrolls still cannot be read, since they have proven impossible to unroll.
Extensive attempts at physical unrolling from the 18\textsuperscript{th} to 20\textsuperscript{th} centuries met with limited success \cite{barker1908book,bics86}, as the scrolls are highly brittle and densely wound.
The sheets are difficult to separate and tend to disintegrate into fragments when disturbed \cite{sider05book}.

\begin{figure}
    \centering
    \includegraphics[width=\linewidth]{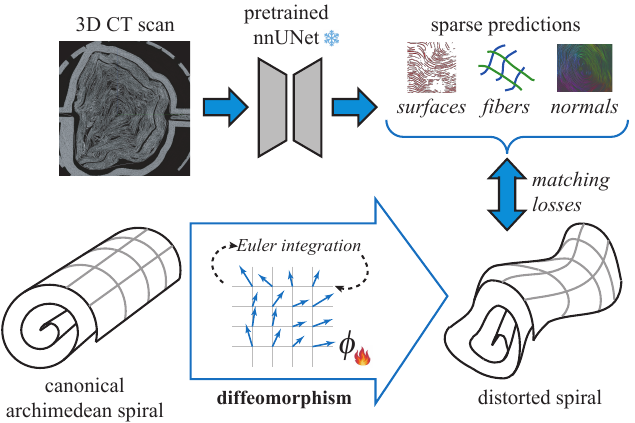}

    \vspace{-6pt}
    \caption{
    Our approach ingests noisy volumetric U-Net predictions of papyrus surface in a CT scan of a rolled scroll, and virtually unrolls it into a continuous 2D sheet surface. It fits a parametric model of an idealized scroll and a diffeomorphic transform given as the integral of a velocity field.
    }
    \label{fig:pipeline}
\end{figure}

\begin{figure*}
\centering
\begin{subfigure}[b]{0.15\linewidth}
    \includegraphics[height=2.62cm]{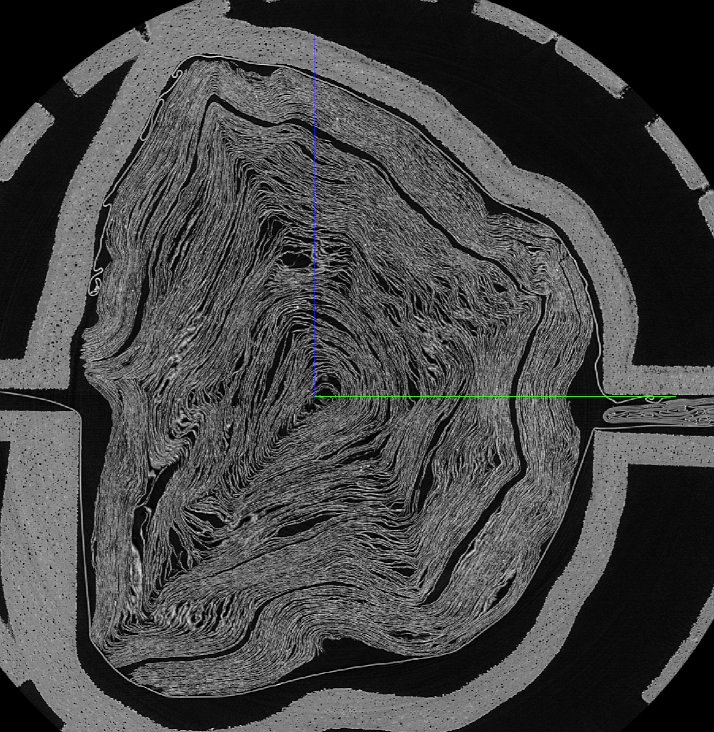}
    \caption{}
    \label{fig:phpf-cross-section}
\end{subfigure}%
\hfill
\begin{subfigure}[b]{0.26\linewidth}
    \includegraphics[height=2.62cm]{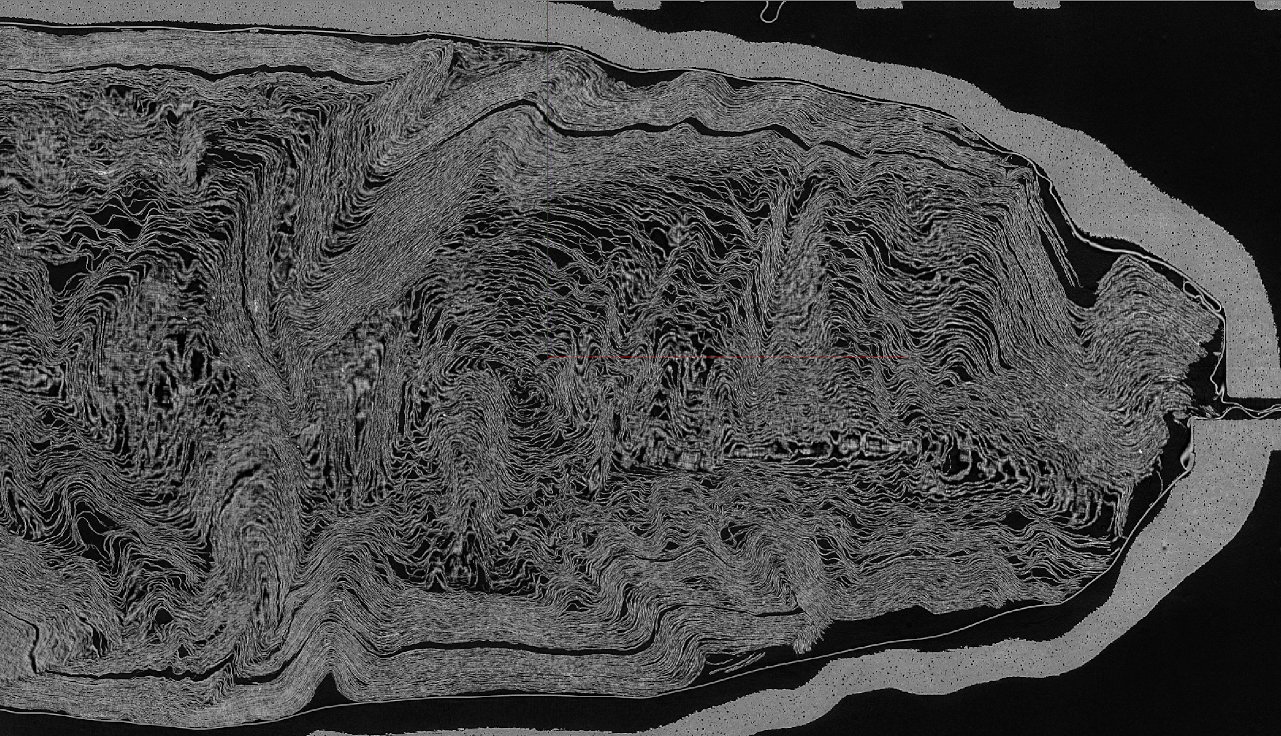}
    \caption{}
    \label{fig:phpf-longwise-section}
\end{subfigure}%
\hfill
\begin{subfigure}[b]{0.18\linewidth}
    \includegraphics[height=2.3cm,trim={0 0 207em 0},clip]{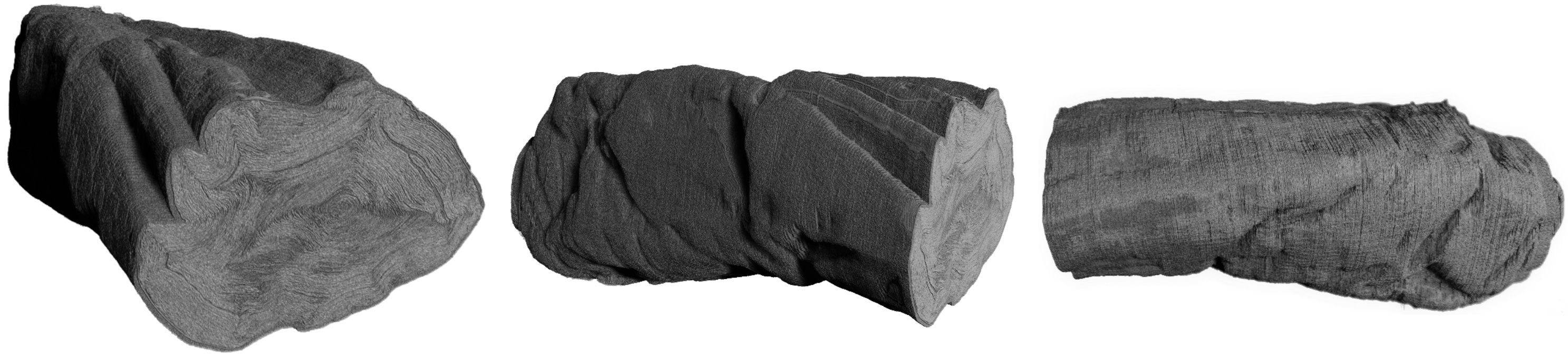}

    \vspace{5pt}
    \caption{}
    \label{fig:phpf-vol-render}
\end{subfigure}%
\hfill
\begin{subfigure}[b]{0.15\linewidth}
    \includegraphics[width=\linewidth]{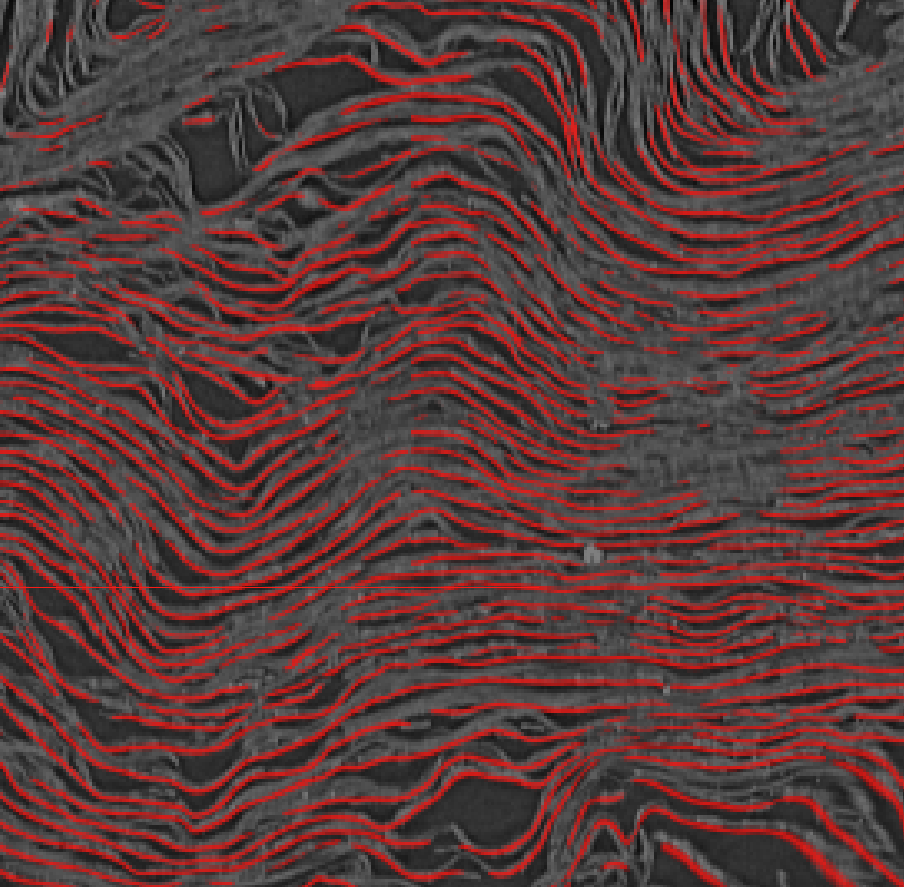}
    \caption{}
    \label{fig:unet-seg-errors}
\end{subfigure}
\hfill
\begin{subfigure}[b]{0.245\linewidth}
    \resizebox{1\linewidth}{!}{%
    \includegraphics[height=2cm]{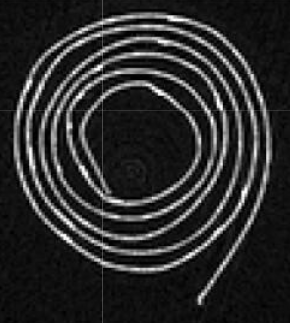}
    \includegraphics[height=2cm]{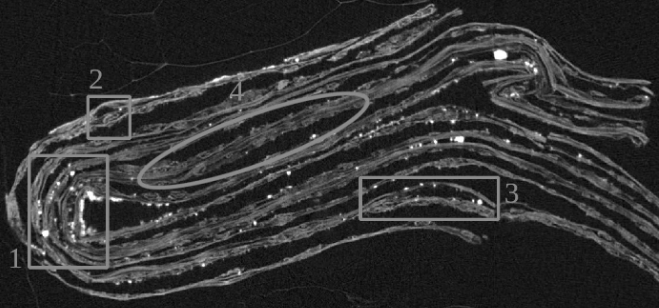}%
    }
    \includegraphics[width=1\linewidth]{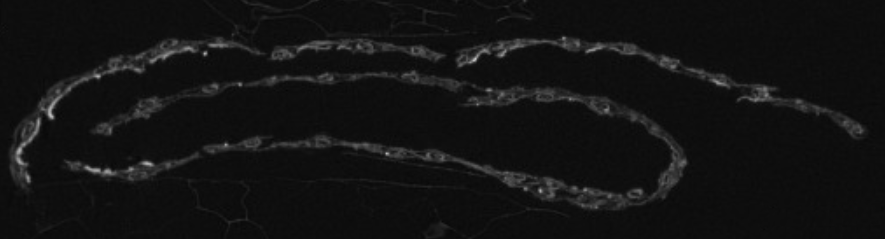}
    \caption{}
    \label{fig:datasets}
\end{subfigure}%

\vspace{-6pt}
\caption{
    \textbf{(a--c)}
    Cross-section, longwise section and volume rendering of \phpf{}, from the micro-CT scan in EduceLab-Scrolls \cite{parsons23dataset}.
    The original scroll is highly compressed and distorted, and it is difficult to discern where the original 2D papyrus surface passes.
    \textbf{(d)}
    State-of-the-art segmentation of surface voxels (red) overlaid on the scroll. Note there are breaks in certain regions, as well as falsely joined sheets and spurious fragments. The predictions do not allow directly extracting a manifold surface.
    \textbf{(e)}
    Representative examples of datasets used in prior works on virtual unrolling/unfolding (CT-OCR-2022 \cite{polevoy22ctocr} used in \cite{kulagin24icdar}; 
    Papyrus L/El227b/3-pU from \cite{klenert25wacvw}; 
    Papryus L/El227b/1-pC from \cite{mahnke20jch}); compare with the complexity of \phpf{} seen in (a--d). 
}
\end{figure*}

One solution to access the text is to acquire 3D micro-CT scans of the rolled scrolls \cite{parsons23dataset} (Fig.~\ref{fig:phpf-cross-section}).
To recover legible text, these 3D scans are \textit{virtually unrolled} \cite{seales04jcdl,seales16sciadv,parsons2020icch,nicolardi24zfpe}.
This involves determining how the original 2D papyrus surface passes through 3D scan volume, typically by manual tracing.
Once this is known, we can extract a thin, flattened volume from the original scan, centered on the original sheet surface.
This enables visual inspection or automated ink detection similarly to a physically-unrolled scroll.

Virtually unrolling the Herculaneum Papyri is highly challenging, since their interior structure is very complex \cite{parsons23dataset}.
The original sheets consisted of two layers of papyrus fibers pressed together \cite{lambert97traces,wallert89papyrus}; however the sheets have frayed and separated over time, leaving it unclear which parts constitute the original inked surface.
Moreover, the scrolls contain hundreds of highly-compressed and distorted windings within a radius of a few centimeters, making it challenging to distinguish one sheet from the next even in high-resolution scans.
Consequently, tracing out the papyrus surface manually is exceedingly time-consuming.

It is therefore desirable to automate the virtual unrolling process.
This requires automatically predicting a 2D manifold surface specifying where the papyrus sheet lies in the 3D scan volume.
This surface is typically represented as a 3D surface mesh, accompanied by a 2D UV coordinate parameterization that defines how to flatten it.
Some works directly use high absorbance voxels to localize the papyrus sheet, followed by heuristics to extract a manifold 2D surface \cite{schilliger24thaumato,seales16sciadv,allegra15eusipco}.
Other approaches instead use neural network predictions of which voxels lie on the surface \cite{klenert25wacvw,kulagin24icdar}.
However even state-of-the-art voxel-based surface predictions for the Herculaneum Papryi do not yield a single clean surface from which a mesh can be `read off'---instead there are many locations where surfaces touch, or are broken.
Moreover, existing methods do not aim to produce a single contiguous surface as output, instead yielding multiple fragments and/or holes; indeed they lack awareness that the scroll originally took this form.

In this work, we present a new method for automated virtual unrolling.
Our key contribution is to leverage prior knowledge of the scrolls' structure to design a model-based approach.
Specifically, we know the original physical scroll was a single rectangular sheet, rolled up into a cylinder, then distorted in some complex way following the eruption.
We aim to invert this process, \ie to find a `good' rolled sheet and deformation that can explain the observed scan data.
The deformation is a spatial transform that warps an idealized rolled scroll (an extruded Archimedean spiral) into the distorted shape we observe in the scans.
We enforce parametrically that the deformation is a diffeomorphism, \ie~a smooth one-to-one mapping.
This means the topological structure of the idealized scroll is preserved by the deformation---it will still be a single, contiguous 2D sheet without holes or intersections, albeit highly distorted.
As input, our approach only requires imperfect neural network predictions of which voxels lie on the papyrus surface or fibers. 
It supports efficient GPU implementation in standard deep learning frameworks, and leverages sparse representations to avoid loading the scroll volume itself into memory.

Our approach differs substantially from existing methods for automated unrolling and unfolding.
It is a top-down, model-based method that optimizes a well-defined objective function globally across the entire scroll; this contrasts with typical bottom-up methods that lack global context, yield fragmentary or non-manifold outputs, and/or are unsuited for the scale and complexity of the Herculaneum Papyri \cite{klenert25wacvw,dambrogio21natcom,schilliger24thaumato}.
Our approach always produces a single continuous rolled surface, that also passes through missing (\eg burned away) regions or those undetected during neural network processing.
Its global model of how a scroll is constituted ensures sheets are interpolated correctly across these gaps.

In summary our primary contribution is
\textbf{a new model-based paradigm for automated virtual unrolling, where an arbitrarily-deformed whole-scroll surface model is fitted to noisy surface predictions}.
As secondary contributions facilitating this, we develop
\textbf{(i)} 
a strategy to convert existing noisy neural network predictions 
to a sparse representation that is robust to errors and efficiently captures the information necessary for surface fitting;
\textbf{(ii)}
a set of geometric losses that result in a computationally tractable optimization problem that largely avoids local optima.

We carefully evaluate our method on the scrolls \phpf{} and \phost{} \cite{parsons23dataset}.
Their challenging structure (Fig.~\ref{fig:phpf-cross-section}) makes these scrolls an excellent test-bed for automated unrolling.
Moreover, a gold-standard manually-created surface mesh is available for 30 windings of \phpf{}---we use this for quantitative evaluation, measuring the benefits of different losses and diffeomorphism parameterizations.
Finally we show our method yields better results than the only existing automated unrolling approach suitable for the Herculaneum Papyri \cite{schilliger24thaumato}.%

\section{Background and Related Work}
\label{sec:related-work}

\textbf{The Herculaneum Papyri} are a collection of papyrus scrolls discovered in the Villa dei Papiri in Herculaneum in 1752 \cite{sider05book,barker1908book}. They are believed to come from a small library in the villa, possibly belonging to the Epicurean philosopher Philodemus \cite{capasso20philodemus}.
During the eruption in \textsc{ad}79 that also destroyed Pompeii, the scrolls were rapidly carbonized by pyroclastic flows \cite{parsons23dataset}, prior to being buried.
Since excavation, around 1800 scrolls or fragments have been catalogued \cite{bics86}; of these approximately 1000 have never been unrolled.

Unfortunately ink is typically not directly visible even in high-resolution micro-CT scans of the Herculaneum Papyri, since they are written in a carbon-based ink that is almost identical in X-ray absorbance to the carbonized papyrus itself \cite{seales13virtual}.
Still, \cite{mocella15natcomm,bukreeva16scirep} showed propagation-based phase-contrast CT can render ink visible, and \cite{parker19plosone,parsons2020icch} showed that ink not visible as light-dark contrast can sometimes still be detected by neural networks.
Indeed, recent work has successfully applied similar methods to virtually-unrolled scans \cite{nicolardi24zfpe,nader23gpw}, detecting significant volumes of legible text based on the ink's effect on the surface texture of the papyrus.

\paragraph{Virtual unrolling.}
The first work to demonstrate the feasibility of virtual unrolling from CT scans was \cite{seales04jcdl}, which applied simple thresholding then meshing, on modern replica papyrus scrolls with fewer than five windings.
Seales \textit{et al.}~\cite{seales16sciadv} virtually unrolled the En-Gedi scroll via a largely automated approach growing a surface out from a seed point; this revealed text in that scroll for the first time.

A scan of the Jerash silver scroll was unrolled using an automated method to extract surfaces from strips \cite{hoffmannbarfod15scirep}, which were re-combined manually. \cite{baum21jerash} instead use Otsu thresholding \cite{otsu79} then skeletonization \cite{fouard06tmi}, and incorporate manual correction.
Otsu thresholding was also applied to a modern papyrus scroll with few windings in \cite{allegra15eusipco}, using region-growing to create the surface mesh.

Baum \textit{et al.}~\cite{baum17apa} extract surfaces of a small papyrus package with around 10 layers, manually tracing the surface in cross-sections, then interpolating to other slices.
A similar approach is taken in \cite{mahnke20jch}, applied to several of the Elephantine Papyri, some of which are folded and badly distorted.
In \cite{dambrogio21natcom} CT scans of folded letters are segmented using \cite{steger98pami} followed by heuristic postprocessing to clean up; \cite{wang22ole} processes terahertz wave scans similarly.
Samko \textit{et al.}~\cite{samko11bmvc,samko14pr} analyze simple parchment scrolls, using graph-cuts to separate the clearly-delineated surfaces; \cite{liu18tip} instead uses \cite{otsu79} then heuristics to separate fused sheets.

Klenert \textit{et al.}~\cite{klenert24tvcg} presents a method for automatically extracting a surface from a binary volume, based on ridge points \cite{eberly94jmi}.
This was applied to a small folded papyrus sheet \cite{mahnke20jch}, as well as a folded silver sheet, showing better performance than \cite{dambrogio21natcom}, and also than other methods for ridge extraction \cite{schultz09tvcg,algarni19pami}.
This is extended in \cite{klenert25wacvw} by replacing the raw CT intensities with predictions from a U-Net \cite{ronneberger15unet,isensee21nnunet} trained to segment the papyrus surface; subsequent processing then follows \cite{klenert24tvcg}.
Similarly, \cite{kulagin24icdar} uses a neural network to find surface voxels, then skeletonization to extract a mesh, with heuristics to address failure cases.
They only evaluate on six documents with simple structures---fewer than ten well-spaced windings and a high-quality unbroken surface everywhere \cite{polevoy22ctocr}.

On the Herculaneum Papyri themselves, there have been fewer efforts.
They are uniquely challenging \cite{seales16sciadv,baum21jerash,polevoy22ctocr}, as they contain hundreds of tightly-packing windings that have been compressed, distorted and sometimes broken following the eruption \cite{parsons23dataset}.
Still, \cite{bukreeva16scirep} consider small regions of two scrolls, while \cite{stabile21scirep} consider two small, nearly-flat fragments; they manually trace out the surface mesh, then flatten it using \cite{levy02tog,sheffer01ewc}.
Nicolardi \textit{et al.}~\cite{nicolardi24zfpe} apply deep learning techniques to detect ink signals in a virtually unrolled part of \phpf{} based on a manually-created mesh (part of the region considered in this work), while \cite{schilling25fasp} presents a semi-automated approach using U-Net surface predictions.
The only existing fully-automated approach that can tackle large portions of Herculaneum Papyri is ThaumatoAnakalyptor \cite{schilliger24thaumato}.
It uses intensity gradients to locate the papyrus sheet and segments this into surface fragments using Mask3D~\cite{schult23icra}.
Heuristics determine whether nearby instances lie on the same winding, with postprocessing to ensure a manifold surface.
This method fails for broken or highly compressed regions, and its heuristics do not optimize any well-defined objective.

None of these methods have any global awareness that the scroll was originally a single sheet rolled or folded, nor do they aim to produce a single surface as output---they yield multiple fragments in challenging cases, and rely on continuity and separability of the surface predictions.

\paragraph{Learning and representing diffeomorphisms.}
One common application of 3D diffeomorphisms is in medical image registration \cite{christensen94,arsigny06,vercauteren09ni,dalca19mia}.
Here we seek a spatial transform that maps anatomically-corresponding points in two 3D scans onto each other.
The mapping is often represented as diffeomorphism since topology should be preserved---there should be no tears or singularities, and the mapping should be injective.
In this work, we instead use a diffeomorphism to transform an idealized spiral-shaped scroll to match a micro-CT scan.
We parameterize it as the integral of a time-constant flow field \cite{ashburner07ni}; however we use explicit Euler integration instead of scaling-and-squaring \cite{moler03siam}, since we found the latter to be unstable. 
Another use of diffeomorphisms in machine learning is normalizing flows \cite{rezende15icml,kingma16neurips,papamakarios21jmlr}, which are flexible learned probability densities, that smoothly and invertibly transform points drawn from an easy-to-sample distribution.
Continuous normalizing flows \cite{chen18neurips,grathwohl18iclr} solve a differential equation defined by a neural flow field, similar to us.
However, we define the field directly on a grid with trilinear interpolation, and use unrolled optimization to backpropagate, rather than the adjoint method.

\section{Dataset}
\label{sec:data}

For this work, we use publicly-available scan data from the EduceLab-Scrolls dataset \cite{parsons23dataset}.
Specifically, we focus on \textit{\phpf{}} and \textit{\phost{}}.
Both scrolls were excavated between 1752 and 1754; 
each measures approximately 15--20cm long by 5cm diameter, and both are heavily distorted (Fig.~\ref{fig:phpf-cross-section}--\ref{fig:phpf-vol-render}).
The X-ray micro-CT scans from \cite{parsons23dataset} were acquired using a synchrotron source at 54keV beam energy with \SI{7.91}{\um} resolution.
The overall sizes are $14376\times7888\times8096$ and $21000\times6700\times9100$ voxels respectively.
These scans are much more complex than those typically considered in virtual unrolling---compare Fig.~\ref{fig:phpf-cross-section} with the examples from recent works in Fig.~\ref{fig:datasets}.
For \phpf{}, we specifically consider a region near the center that has been the focus of recent manual unrolling attempts \cite{nicolardi24zfpe,nader23gpw}; this subvolume has a bounding-box of size $13600\times4200\times3600$ voxels.
A ground-truth surface mesh is available representing the geometry of the papyrus sheet in this region, enabling us to conduct quantitative experiments.
This covers approximately 30 windings, and was was created by a \textit{tour de force} manual annotation effort requiring hundreds of hours of human input \cite{parkerXXvolcart}.

\section{Feature Extraction and Postprocessing}
\label{sec:features}

Given a CT scan of a scroll, we aim to extract a compact representation containing salient features such as probable locations of the sheet surface and its normals, which will be used to guide the automated unrolling process.

\paragraph{Neural network predictions.}
We use three existing publicly-available volumetric U-Net \cite{ronneberger15unet} segmentation models, trained using nnUNet-v2~\cite{isensee21nnunet} on a tiny fraction of \textit{\phpf{}}\footnote{models and predictions obtained from \url{https://dl.ash2txt.org/community-uploads/bruniss/scrolls/s1/}}.
They output probabilities that each voxel belongs to (i) the surface of the papyrus (similar to \cite{klenert25wacvw}); or (ii) a horizontal or vertical fiber.
We emphasize that these segmentation tasks are very challenging---although the CT scan reveals papyrus as bright against a dark background, sheets are often tightly packed together, highly curved, or even broken.
Sometimes individual strips of papyrus have frayed away from the surface; sometimes the entire front and back surfaces have separated over a larger area.
Consequently, even these state-of-the models have errors in their predictions,
e.g.~cases where sheets or fibers are falsely joined, or entirely missing from the segmentation (see Fig.~\ref{fig:unet-seg-errors}).
In this work, our focus is on extracting a plausible 2D manifold surface from these predictions despite their imperfections.

\paragraph{Feature postprocessing.}
Given the surface and fiber U-Net predictions, our first contribution is to convert these to a more compact representation in a way that is robust to errors, prior to fitting global model of the scroll surface to them.
Specifically, we aim to find sequences of 3D points that should lie on the same winding, \ie adjacent to each other in the unrolled scroll; we term such series of points \textit{paths}.
Our strategy is different for fibers and surfaces, due to the different geometric structures they represent.

\paragraph{Fiber paths.}
We threshold the fiber prediction probabilities at 50\% confidence, and extract connected components (CCs) from the resulting binary volume using the method of \cite{rosenfeld66cc,wu05cc} as implemented in \cite{silversmith21cc3d}. Each connected component is then skeletonized using \cite{silversmith21kimi,sato00teasar}.
Ideally each skeleton would have the form of a chain graph, tracing the midline of the respective fiber, which should form a path lying on the papyrus sheet.
However, the CCs do not necessarily capture individual fibers, as fibers sometimes touch, either physically in the scroll or due to segmentation errors; this results in skeletons with a branching structure.
We therefore further process each extracted skeleton, finding the longest chain between two end-/branch-points, saving this as a fiber path, removing it from the skeleton, and repeating.
This ensures that each fiber path represents only a single fiber belonging to a single sheet, at the expense of sometimes breaking fibers into multiple sections.

\paragraph{Surface paths.}
After thresholding the surface predictions, we observe that na\"ive use of 3D connected components yields CCs with many sheets combined, since their segmentation masks touch.
Instead, we extract 2D slices from the volume, and find (2D) CCs within these slices only. Since it is exponentially less probable that two sheets are wrongly joined in one specific slice than globally, this yields CCs that only very rarely include multiple sheets, albeit at the expense of these CCs being only 1D slices of the true 2D surface.
We use the same subsequent processing as for fibers, \ie skeletonization then iterated longest-path extraction.
We repeat this process using slices along each of the canonical planes independently and merge the results.

\paragraph{Surface normals.}
We also extract approximate surface normals, by convolving the surface prediction probability volume with a 3D Sobel filter, then taking a mean within windows of normals whose magnitude exceeds a threshold.
Window centers are chosen by stratified sampling, iterating over grid cells, and sampling points on paths in each cell.

\paragraph{Relative winding numbers.}
For each extracted surface path, we also determine what nearby paths lies on the next winding outwards; this allows us to derive an estimate of the winding spacing.
We cast rays outwards along the local normals from each point along the path, and record the first path hit.
Since there may be breaks in surfaces or spurious fragments, we then construct a graph with one node per surface path, and edges connecting those which were found to be neighboring during the ray-casting process.
We drop nodes that are part of cycles (since these imply a contradiction), then use majority voting by edges to determine whether two paths in fact should be treated as adjacent.

\section{Spiral Fitting}
\label{sec:spiral-fitting}

Our goal now is to fit a model of the deformed scroll to the postprocessed U-Net outputs described above.
This model is an explicit representation of the scroll, as a single, highly distorted, rolled-up surface.
The predicted fiber and surface paths, normals, and relative winding numbers specify a set of constraints on where the surface lies in the scan volume.
We set up a continuous global optimization problem of finding the geometric parameters of the original rolled scroll, and a smooth, invertible deformation (a diffeomorphism) that transforms it to its distorted state in the scan, aiming to satisfy the constraints as well as possible.

In Sec.~\ref{sec:method/canonical-spiral}, we describe the idealized form of the non-deformed scroll in a canonical space, and its parameters. Then, Sec.~\ref{sec:method/transform-parameterisation} discusses the diffeomorphism that transforms the idealized scroll to the distorted version seen in the scan volume.
Sec.~\ref{sec:method/losses} describes the losses used to fit the transform to the features from Sec.~\ref{sec:features}.
Lastly, Sec.~\ref{sec:method/meshing} explains how we extract a flattened surface mesh.

\begin{figure*}
\vspace{-4pt}
\includegraphics[width=0.30\linewidth]{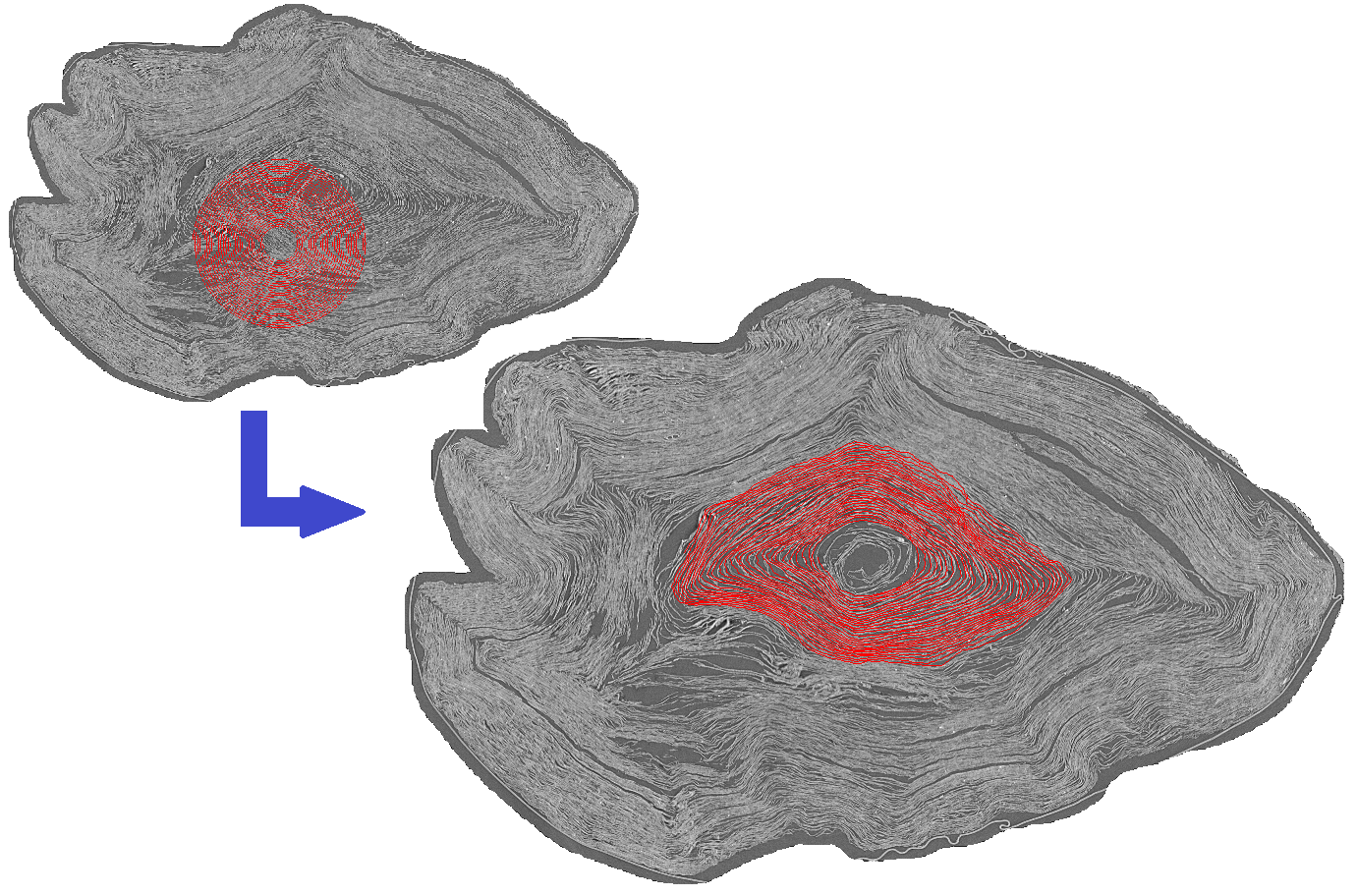}%
\includegraphics[width=0.21\linewidth]{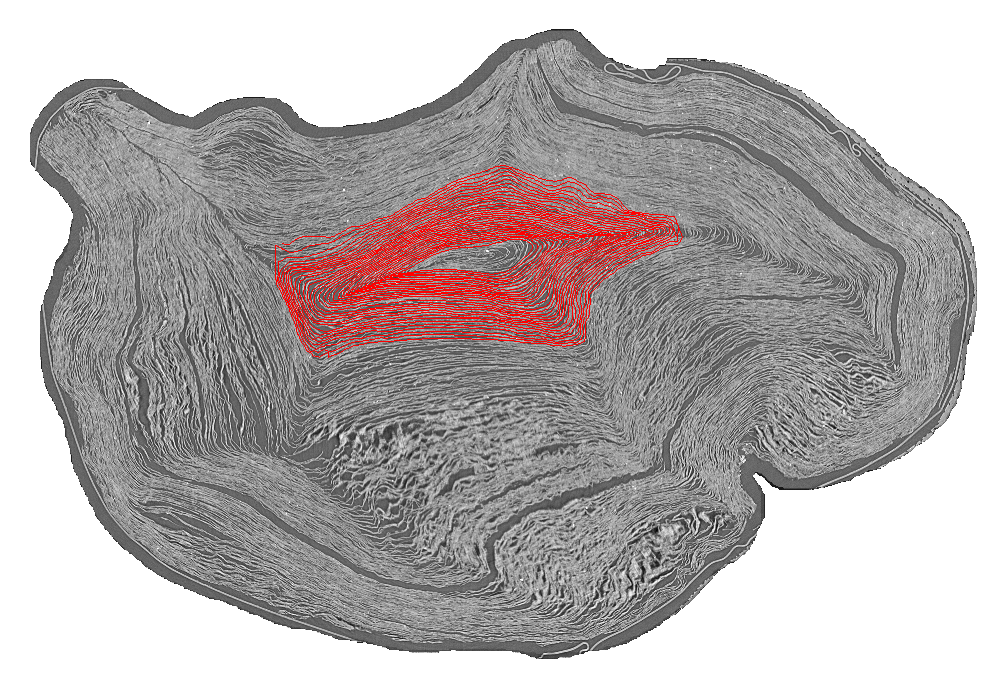}%
\includegraphics[width=0.18\linewidth]{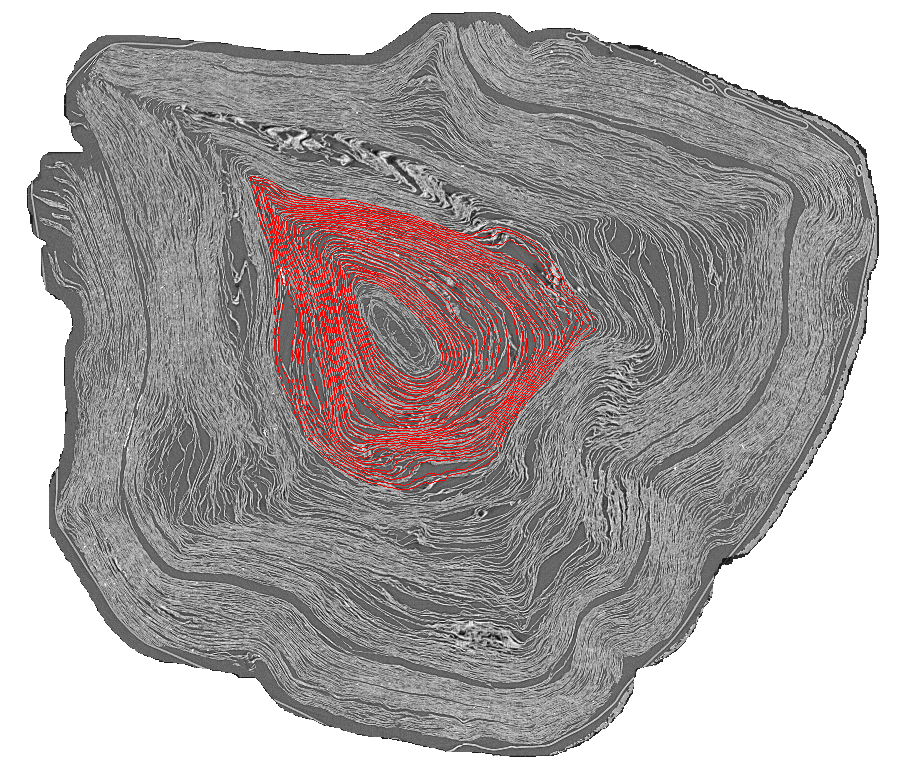}%
\includegraphics[width=0.16\linewidth]{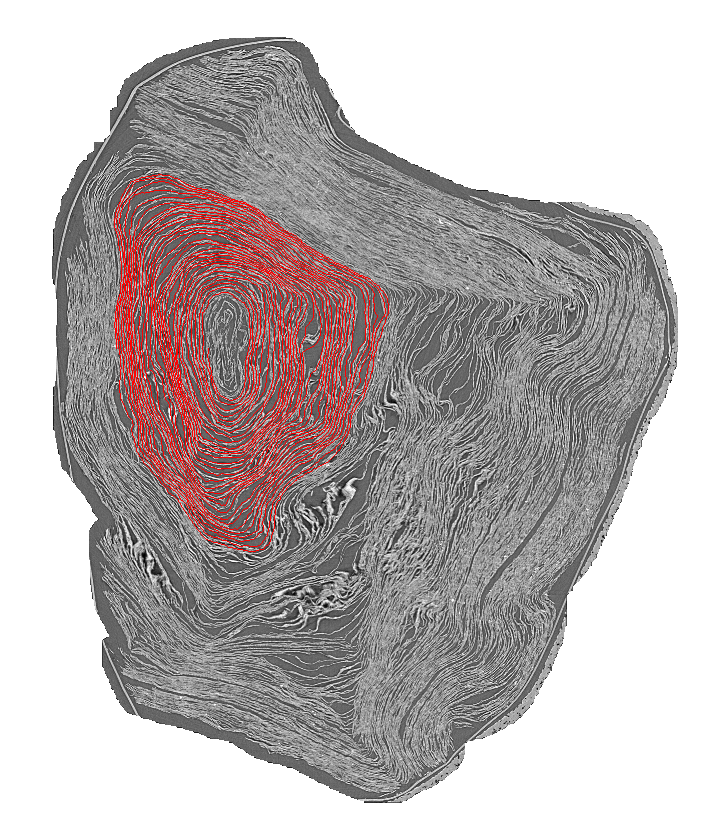}%
\raisebox{6pt}{\includegraphics[width=0.16\linewidth]{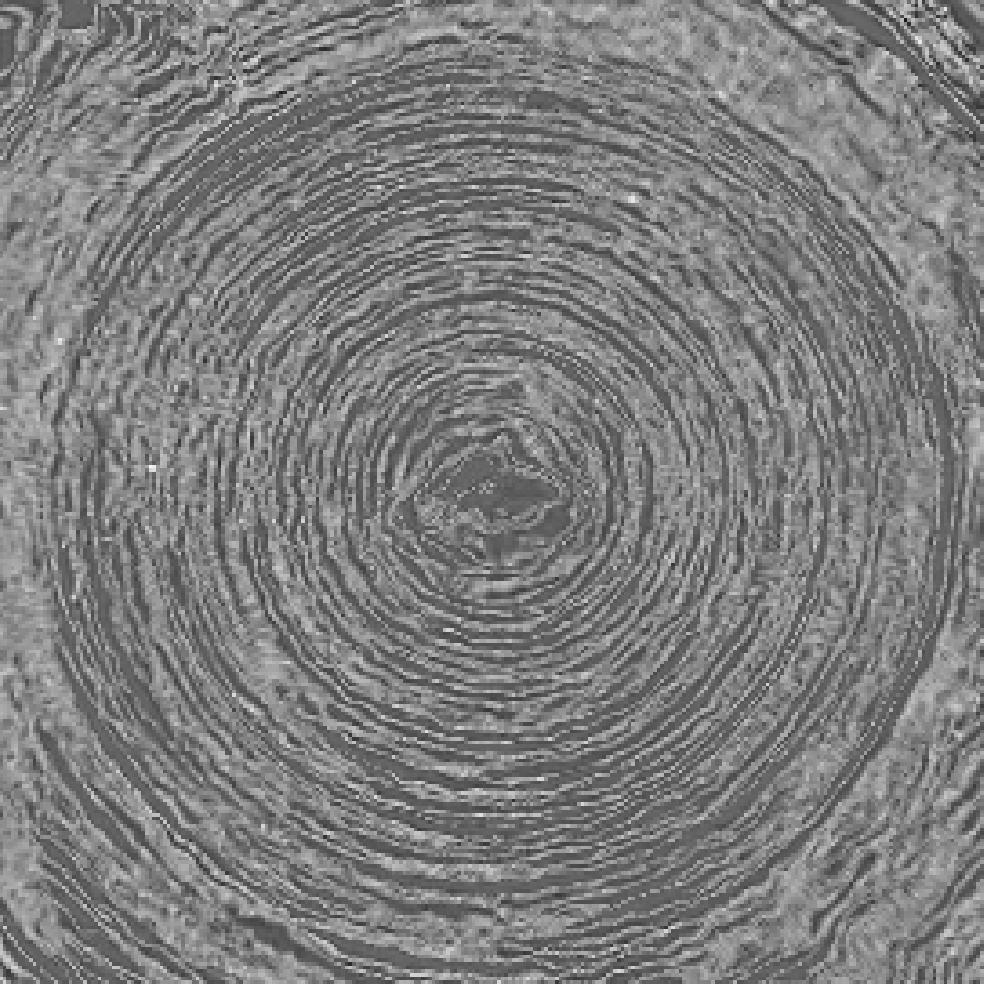}}
    
    \vspace{-8pt}
    \caption{
    Cross-sections of \phpf{} with the spiral overlaid in red, before optimization when the diffeomorphism is an identity transform (top-left inset); and after optimization. During fitting the spiral adapts to follow the distorted scroll surface.
    Far right: Undeformed cross-section, given by transforming the scan with the inverse diffeomorphism.
    The distorted windings become near-circular.
    }
    \label{fig:s1-sections}
    
\end{figure*}

\subsection{Canonical spiral representation}
\label{sec:method/canonical-spiral}

We define a `canonical' pre-deformation space, in which the scroll is assumed to have taken the form of an Archimedean spiral in cross-section; this is extruded perpendicular to the plane to create a 2D surface.
The spiral cross-section of the canonical scroll has only one unknown parameter $\omega$, how tightly it winds (\ie~the rate of change of radius wrt angle). We assume the direction of winding (clockwise vs anticlockwise) is known, since this is easy to observe visually.
We also assume the canonical spiral is centered in the $xy$-plane and extruded along the $z$-axis.
Thus, 3D points $s$ on the canonical 2D surface are given by
\begin{equation}
s(\theta ,\, z ;\, \omega) =
\left(
\begin{array}{c}
     \frac{\theta}{\omega} \cos \theta \\
    -\frac{\theta}{\omega} \sin \theta \\ 
    z
\end{array}
\right)
\; \forall \, \theta > 0 ,\, z_\mathrm{min} < z < z_\mathrm{max}
\label{eq:spiral}
\end{equation}
While this representation is naturally somewhat idealized
(for example the real papyrus sheet has a non-infinitessimal thickness),
it still reflects the overall structure of the original rolled scroll.

\subsection{Transform parameterization}
\label{sec:method/transform-parameterisation}

We next describe a multi-stage spatial transform $T_{S\rightarrow V}$ mapping from the canonical space of the original spiral-shaped scroll, to the 3D space of the scan volume.
This transform models the deformation of the perfect canonical scroll into the complex shape we observe in the scan.
We enforce parametrically that the deformation is a diffeomorphism, \ie~a smooth one-to-one mapping. This means the topological structure of the canonical scroll is preserved during distortion---it will still be a single, contiguous sheet after deformation.

Specifically, the transform is a composition $T_{S\rightarrow V} = T_\mathrm{aff} \,\circ\, T_\mathrm{flow} \,\circ\, T_\mathrm{gap}$, where:
(i) $T_\mathrm{aff}$ is a per-slice affine (scaling and translation) transform;
(ii) $T_\mathrm{flow}$ is the result of integrating a flow field;
(iii) $T_\mathrm{gap}$ rescales inter-winding gaps.
We describe each of these stages below, and provide full  mathematical details in the supplementary material.

\paragraph{Per-slice scale and translation.}
We non-isotropically scale and translate each slice in the $xy$-plane; the scale factors and translations vary wrt $z$.
This stage is parameterized by $N \times 2$ log-scales where $N$ is a fixed number of keypoints along the $z$-axis; the translations and log-scales are linearly interpolated for other $z$ values.

\paragraph{Integrated flow field.}
We introduce a flow field $\mathbf{u}$, \ie a 3D vector field of 3D velocities over the scan volume.
This allows us to transform a given point by simulating its movement in the flow for a fixed length of time.
This is done by solving an ODE defined by $\mathbf{u}$ and an initial condition (the starting point).
Writing $\phi^{(t)}$ for the transform induced by simulating the flow $\mathbf{u}$ for time $t$,
the final transformed position of a point $\mathbf{x}$ is given by $T_\mathrm{flow}(\mathbf{x}) = \phi^{(1)}(\mathbf{x})$, where $\phi^{(t)}$ evolves according to
\begin{equation}
    \frac{d}{dt}\phi^{(t)}(\mathbf{x}) =
    \mathbf{u}\left(
        \phi^{(t)}(\mathbf{x})
    \right)
\end{equation}
with the initial condition $\phi^{(0)}(\mathbf{x}) = \mathbf{x}$, \ie we start with an identity transform.
The resulting transformation is guaranteed to be a diffeomorphism, provided the flow field $\mathbf{u}$ is itself smooth \cite{ashburner07ni,arsigny06}.
Formally, the space of flow fields is the Lie algebra that generates the Lie group of diffeomorphisms; the former provides a straightforward way to parameterize the latter.
Moreover, the inverse diffeomorphism is easily found by negating the flow field, then integrating backwards in time.
We parameterize the flow field by trilinear interpolation of coarse and fine grids of 3D velocity vectors defined over the scan volume, where the fine grid is $48\times$ lower resolution than the raw scan, and the coarse grid $6\times$ lower again.
We solve the ODE using explicit Euler integration with 16 steps.
Note that in principle this stage is sufficiently expressive without the other stages; however we find that the others ease optimization (by allowing more global updates), and reduce the required resolution (hence memory cost) of the flow field.

\paragraph{Gap scaling.}
We define a scalar field over the 2D surface of the canonical scroll spiral, which specifies for each point, by how much the next winding should be pushed outward from its default distance (defined by being an Archimedean spiral), \ie locally scaling the gaps between windings.
This field is parameterized by bilinear interpolation of a 2D grid of log-scales defined over the surface.
Since it is quadratic in memory cost, this provides a more efficient way to gain spatial resolution than increasing the resolution of the 3D grid defining the flow field above.

The overall composed transform $T_\mathrm{S \rightarrow V}$ is diffeomorphic and efficiently invertible (since each stage is); we write $T_\mathrm{V \rightarrow S} = T_\mathrm{S \rightarrow V}^{-1}$. 
This enables defining losses (which requires transforming from scan to canonical space) and performing reconstruction or reversing the distortion (which require transforming from canonical to scan space).
In the next section, we discuss the losses we use to fit the transform parameters, \ie the per-slice log-scales, the 3D flow field, and the 2D gap-scaling field.

\subsection{Losses and optimization}
\label{sec:method/losses}

\newcommand{\loss}[1]{\ensuremath{\mathcal{L}_\mathrm{#1}}}
\newcommand{\TVS}{T_\mathrm{V \rightarrow S}}
\newcommand{\TSV}{T_\mathrm{S \rightarrow V}}
\newcommand{\xV}{\mathbf{x}^\mathrm{V}}
\newcommand{\xS}{\mathbf{x}^\mathrm{S}}

To optimize the parameters of the spiral and the deformation to fit the surface and fiber predictions, we define several losses based on the postprocessed features.

First, \loss{normal} requires that normals of the deformed spiral match those of the surface predictions. For a point $\xV$ in the scan volume, we use the approximate normal $\mathbf{n}$ from the U-Net surface predictions (Sec.~\ref{sec:features}), then calculate
$\xS = \TVS(\xV)$, and the corresponding canonical-space radial direction vector $\boldsymbol{\delta r}=(\xS_0 / ||\xS||,\, \xS_1 / ||\xS||,\, 0)$.
Then 
\begin{equation}
\loss{normal} = 1 - \mathrm{sim}\!\left( \boldsymbol{\delta r} ,\, \TVS(\xV + \mathbf{n}) \right)
\end{equation}
where $\mathrm{sim}$ denotes cosine similarity of vectors.

Second, \loss{radius} requires that paths (\ie series of points on the same surface or fiber, extracted per Sec.~\ref{sec:features}) lie at constant radius in the canonical space, after adjusting for radius change due to the spiral.
For a set of points $\mathcal{P}$ representing such a path, 
\begin{equation}
\loss{radius} = \frac{1}{|\mathcal{P}|} \sum_{\xV \in \mathcal{P}} |r(\xV) - \langle r \rangle|
\end{equation}
where 
$
r(\xV) = ||\xS-\xS\,{\cdot}\,(0,0,1))|| - \omega\arctan\!\left( \xS_2/\xS_1 \right),
$
$\xS = \TVS(\xV)$, and
$\langle r \rangle = \frac{1}{|\mathcal{P}|} \sum_{\xV \in \mathcal{P}} r(\xV)$.

Third, \loss{windings} requires that estimated relative winding numbers between surfaces should be respected, \ie if two points in the scan volume are predicted to be $K$ windings apart, then they should also be $K$ spiral windings apart when mapped to canonical space.

The above losses are designed to be robust to local optima; this is important due to the repetitive layered structure of the tightly-wound papyrus sheets.
The losses aim to match the orientation and density of scroll windings, but not the exact `phase' or alignment of individual sheets along the normal direction.
This means the model can freely shift windings outwards or inwards to match the overall structure of the scroll, avoiding local optima.

However, we also aim for precise alignment of our deformed spiral with individual sheets, not just correct spacing and orientation.
Therefore, during the second half of the optimization process, after the overall structure is found, we add an extra loss.
\loss{distance} requires each predicted surface and fiber path to be exactly aligned with a winding of our deformed spiral.
For each path $\mathcal{P}$ we find the spiral winding that is nearest to it, and impose an L1 loss between all points of the path and the corresponding nearest points on the deformed spiral winding:
\begin{equation}
\loss{distance} = \frac{1}{|\mathcal{P}|} \sum_{\xV \in \mathcal{P}} \left| r(\xV) - r^*(\xV) \right|
\end{equation}
where
\begin{equation}
r^*(\xV) = 
\begin{cases}
    r(\xV) - m & \text{if $m < \pi\,\omega$} \\
    r(\xV) + 2\pi\,\omega - m & \text{if $m \geq \pi\,\omega$}
\end{cases}
\end{equation}
and
$ m = r(\xV) \,\mathrm{mod}\, (2\pi\,\omega)$.
Note that per (\ref{eq:spiral}), $2\pi\,\omega$ corresponds to the winding spacing in the canonical space.

To ensure that the surface is correctly orientated when flattened, we include two directional losses based on the fiber predictions.
\loss{horizontal} requires horizontal fiber paths $\mathcal{P}$ have constant $z$-coordinate in the canonical space:
\begin{equation}
    \loss{horizontal} = \frac{1}{|\mathcal{P}|} \sum_{z \in \mathcal{Z}} |z - \langle z \rangle|
\end{equation}
where
$\mathcal{Z} = \left\{ \TVS(\xV) \,{\cdot}\, (0,0,1) \,\big|\, \xV \in \mathcal{P} \right\}$.
\loss{vertical} requires vertical fiber paths have constant angular coordinate; it is defined analogously to \loss{horizontal}.

We also include two regularization losses.
\loss{stretch} discourages the deformation from stretching the surface tangentially, so it only bends, or sheets separate: $\loss{stretch} = \big|\,||\TVS(\xV + \boldsymbol{\Delta}) - \TVS(\xV)|| - 1 \big|$ where $\boldsymbol{\Delta}$ is a random unit vector perpendicular to $\mathbf{n}(\xV)$.
\loss{center} encourages the centerline of the scroll to remain at its original position.

Our overall loss is a weighted sum of those discussed above, calculated over minibatches of randomly sampled paths\footnote{further hyperparameter details are given in the supplementary material}.
We minimize this loss wrt the diffeomorphism and spiral parameters using Adam~\cite{kingma15adam}, with a fixed learning rate of $5\times10^{-4}$.
We run for 20K iterations, by which point the overall loss has converged.

\begin{figure*}
    \centering
    \includegraphics[width=\linewidth]{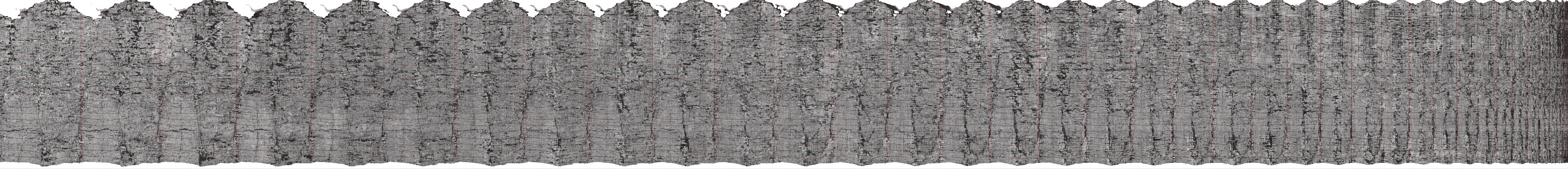}

    \vspace{2pt}
    \includegraphics[width=\linewidth]{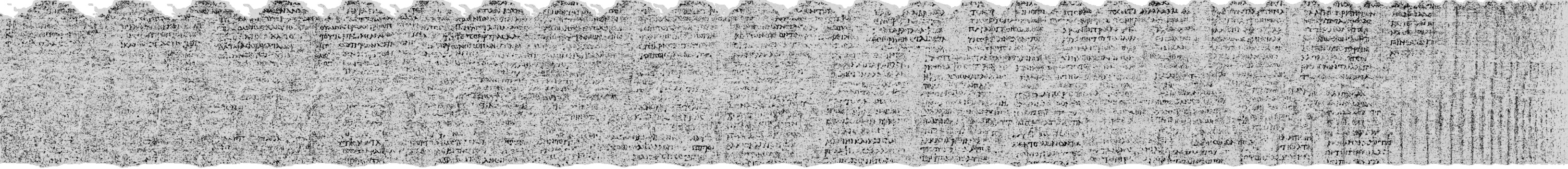}

    \vspace{-5pt}
    \caption{
    \textbf{Top:} Visualization of the papyrus surface extracted from \phpf{} by our method.
    The right-hand end corresponds to the center of the rolled scroll; the wavy edge at the top reflects the increasing radius of windings further from the center.
    \textbf{Bottom:} Ink predictions using the TimeSformer~\cite{bertasius21icml} model from \cite{nader23gpw}, applied to the virtually unrolled volume. Note the rows of Greek characters (black) visible throughout the extracted surface.
    Best viewed with zoom.
    }
    \label{fig:render-and-ink}
    \vspace{-4pt}
\end{figure*}

\begin{figure}
\centering
    \includegraphics[width=0.32\linewidth]{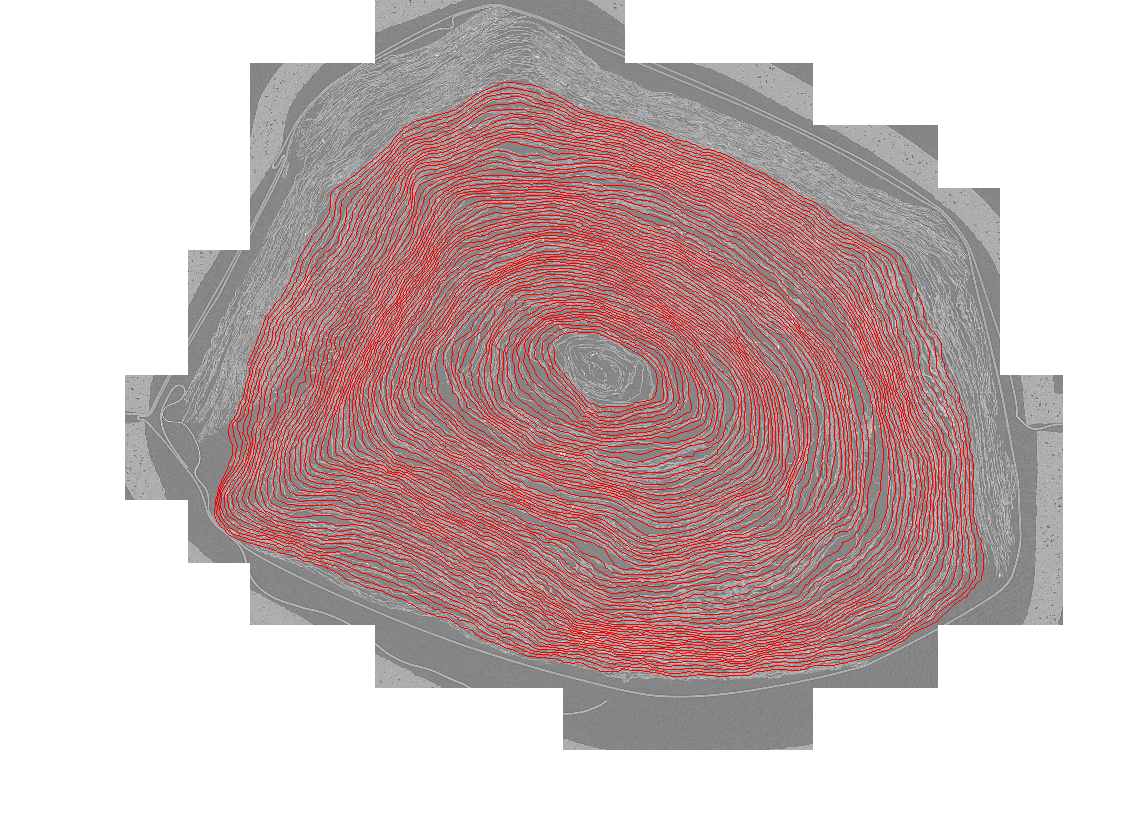}\hfill%
    \includegraphics[width=0.32\linewidth]{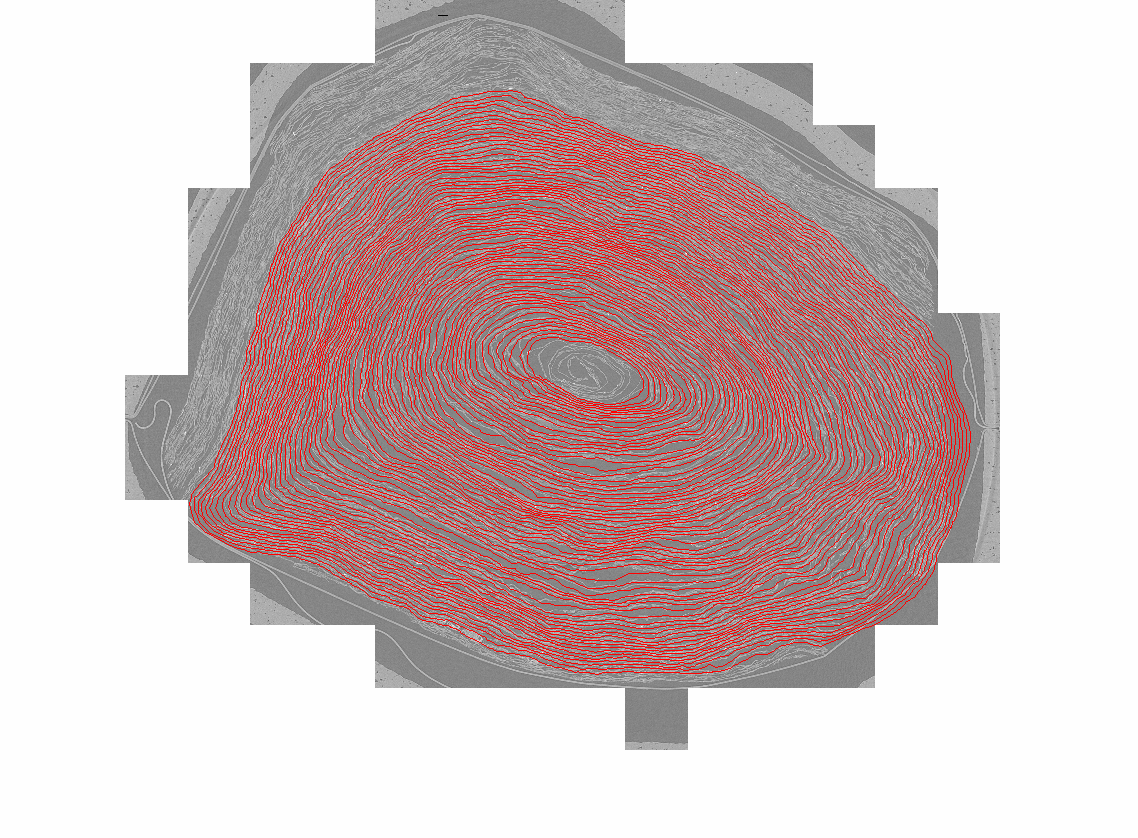}\hfill%
    \includegraphics[width=0.32\linewidth]{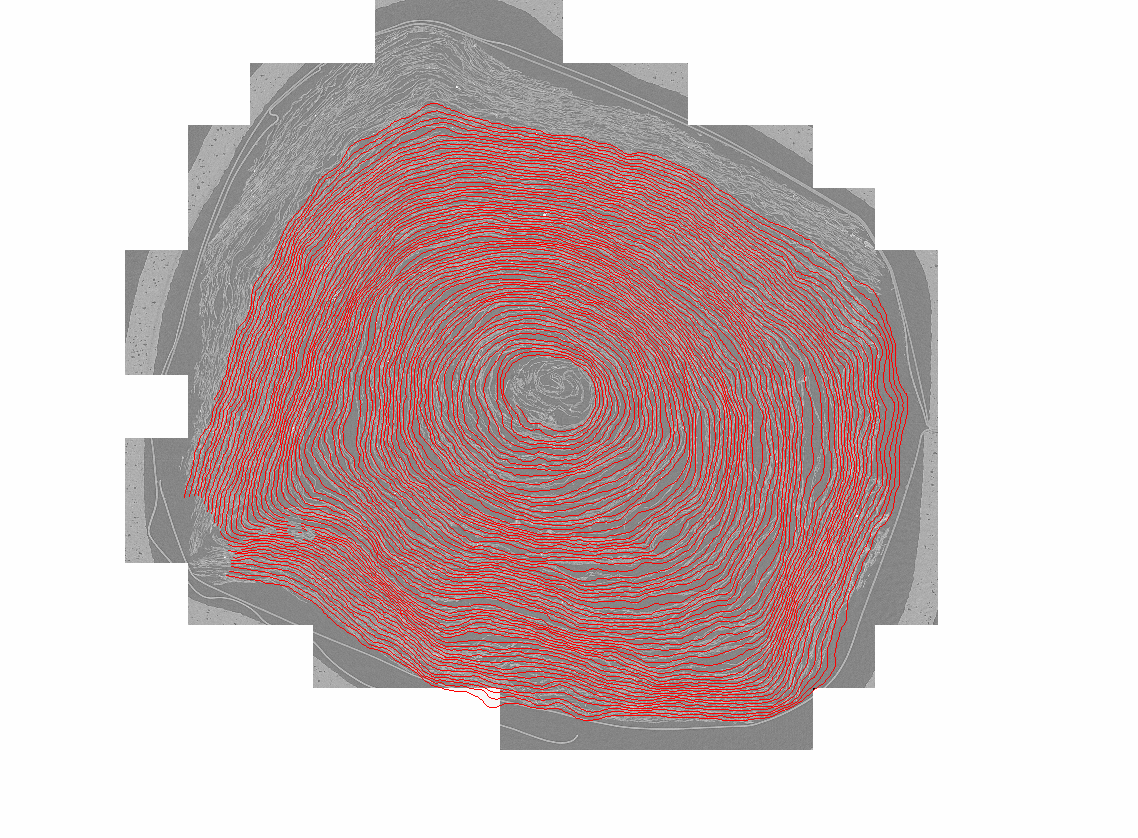}
    
    \vspace{-3pt}
    \resizebox{\linewidth}{!}{%
    \includegraphics[height=2.5cm]{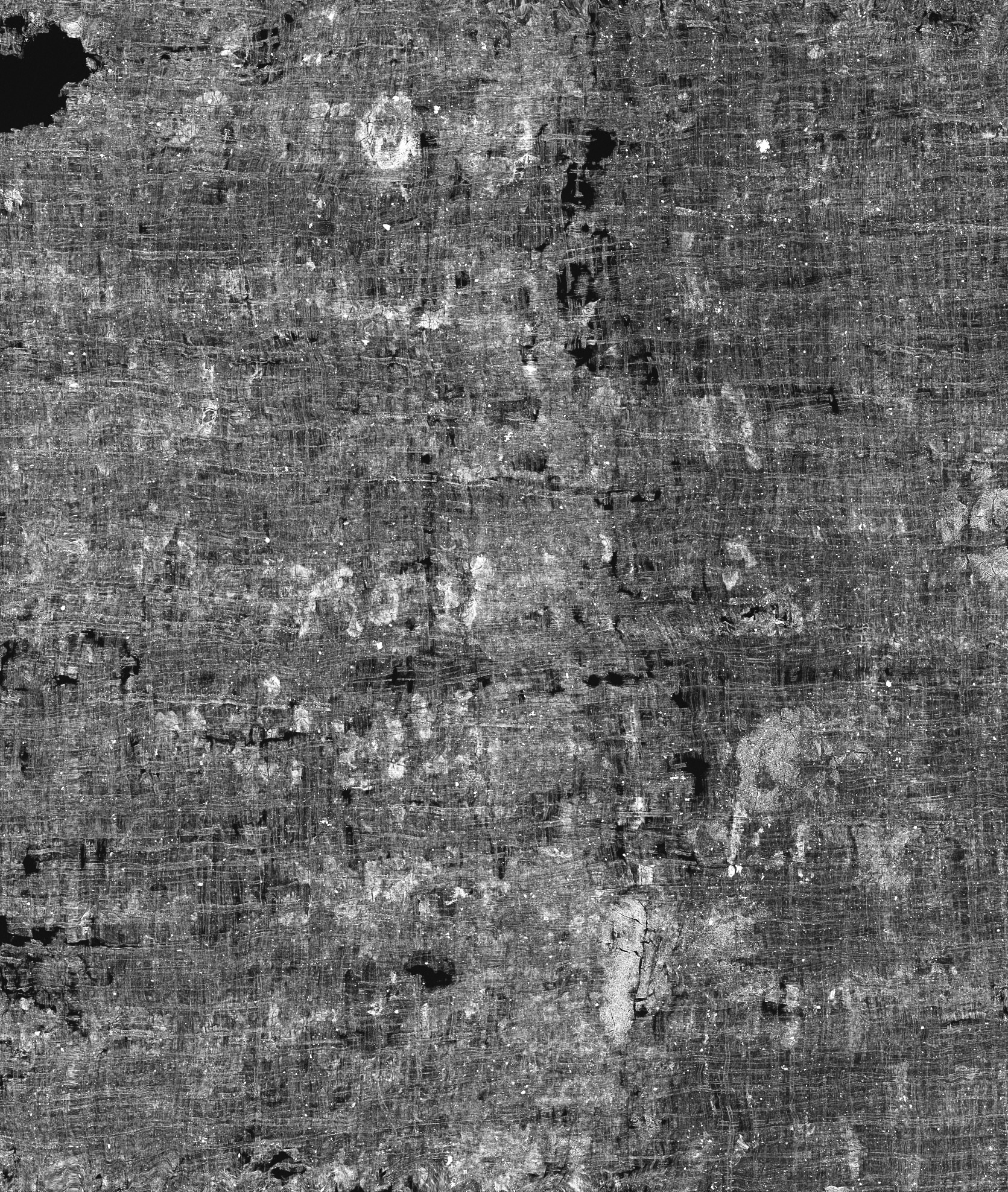}
    \includegraphics[height=2.5cm]{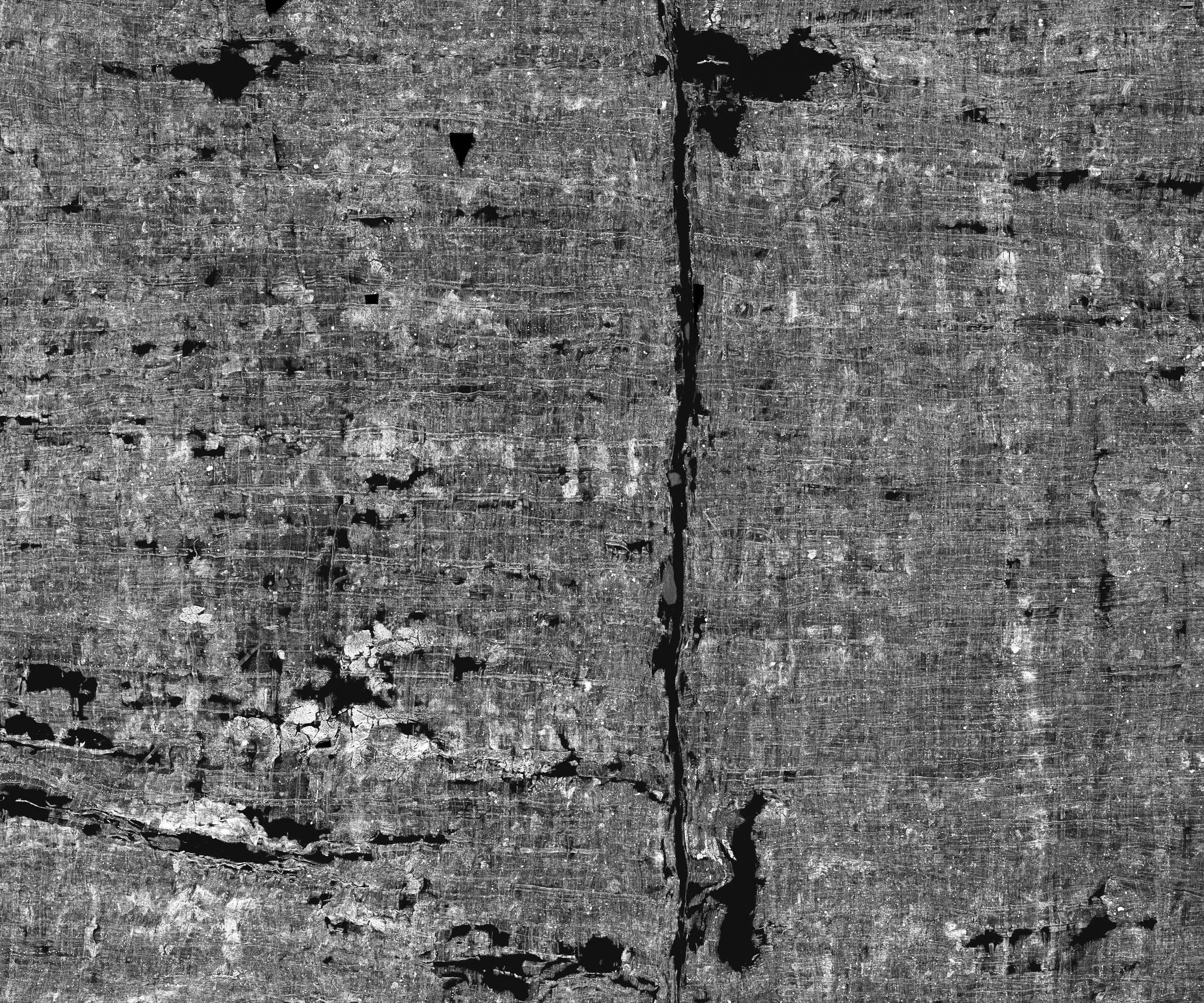}
    \includegraphics[height=2.5cm]{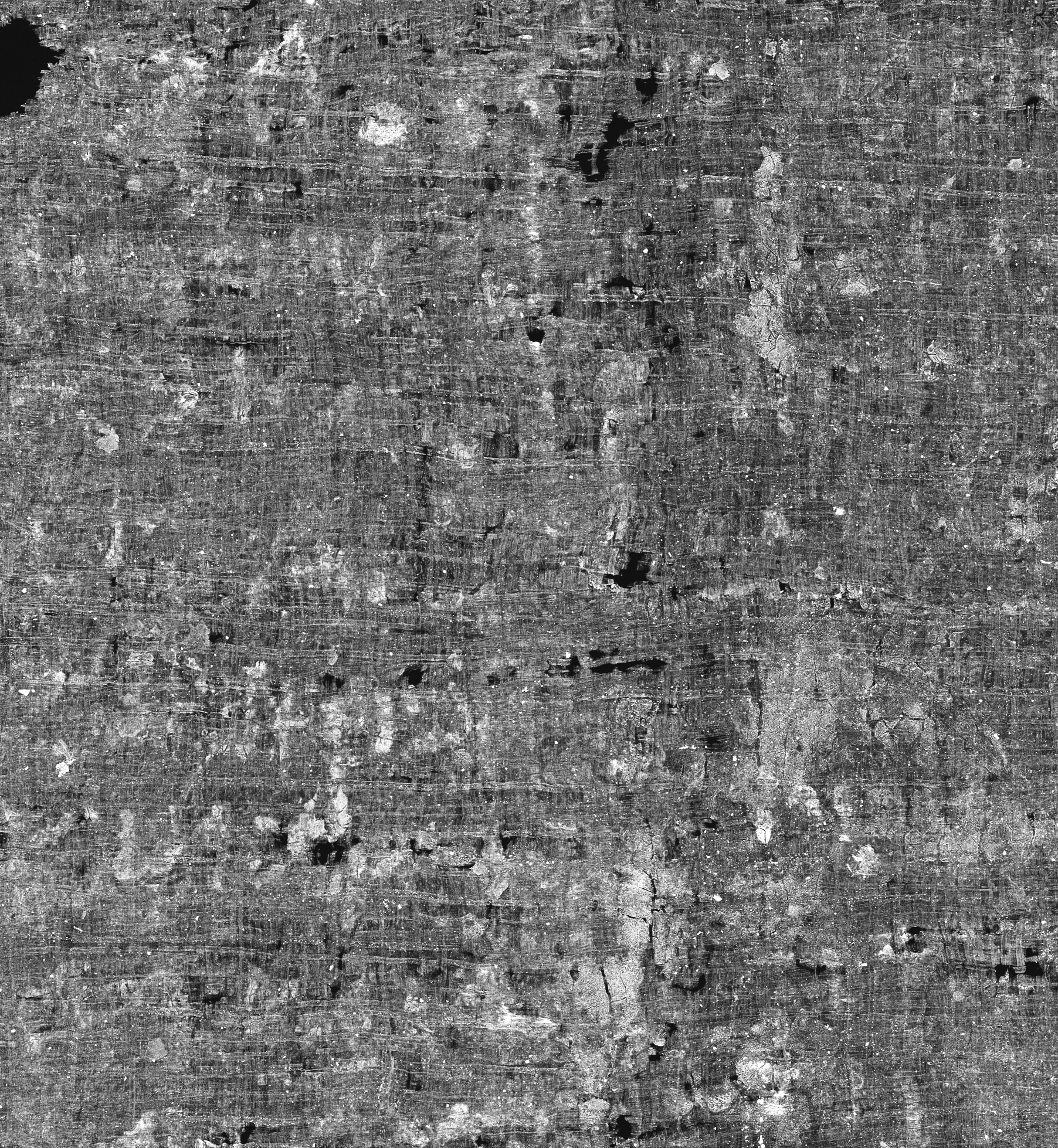}}

    \vspace{-8pt}
    \caption{\textbf{Top:} deformed spiral overlaid on \phost{}. \textbf{Bottom:} Close-ups of the unrolled surface. Ink is visible as light areas; several Greek letters can be discerned. The full output of our method is in the appendix.}
    \label{fig:s5-results}
\end{figure}

\begin{table}[t]
    \centering
    \resizebox{\linewidth}{!}{%
    \begin{tabular}{@{}lccccc@{}}
    \toprule
        ~ & WJF & MRWD & ChD & AD & Str \\
        \midrule
        Ours (full) & 3.20\% & 7.64 & 5.29 & 0.0568 & 1.239 \\ 
        + SLIM & ~ & ~ & ~ & 0.0933 & 1.031 \\
        no surface-normals & 3.08\% & 8.41 & 5.75 & 0.0546 & 1.223 \\ 
        no rel.~wind.~num & 2.77\% & 25.67 & 7.72 & 0.0556 & 1.166 \\ 
        no dist. loss & 3.90\% & 7.46 & 5.93 & 0.0535 & 1.172 \\ 
    \midrule
        no fiber orient.~loss & 3.15\% & 7.51 & 5.36 & 0.0583 & 1.292 \\ 
         + SLIM & ~ & ~ & ~ & 0.0933 & 1.013 \\
        no fiber losses & 3.75\% & 6.83 & 5.46 & 0.0533 & 1.214 \\ 
         + SLIM & ~ & ~ & ~ & 0.0925 & 1.005 \\
    \midrule
        ThaumatoAnakalyptor & -- & -- & 5.59 & 0.1567 & 1.027 \\
    \bottomrule
    \end{tabular}
}
    \vspace{-6pt}
    \caption{Comparison of several variants and ablations of our approach, and the previous state-of-the-art automated unrolling method ThaumatoAnakalyptor. `Ours (full)' uses only fibers for UV embedding (flattening); `+ SLIM' uses \cite{rabinovich17tog} for a post-hoc flattening (UV-embedding) stage. Lower is better for all metrics}
    \label{tab:loss-ablations}
\end{table}

\subsection{Extracting and flattening a mesh}
\label{sec:method/meshing}

After fitting the diffeomorphism and spiral parameters, it is desirable to extract the papryus surface as a mesh.
In the canonical spiral space, we can directly create a quad-mesh by building a lattice of points sampled with regular spacing $\delta\theta$ around the spiral (from the center out to the edge) and $\delta z$ along its length.
We then transform the vertices of this mesh to the scan volume using the diffeomorphism.
Thus, the $(i,\,j)$\textsuperscript{th} vertex of the final quad-mesh is given by $v_{ij} = \TSV\left(s(i\,\delta\theta,\,j\,\delta z;\,\omega)\right)$
The resulting mesh is guaranteed to be manifold and free of intersections, since it is given by applying the (topology-preserving) diffeomorphism to the spiral surface that already has these characteristics.

In order to display the flattened surface, we must also associate a 2D UV coordinate with each vertex of the mesh, defining where in the plane it is placed after flattening.
Typically \cite{stabile21scirep,baum21jerash,schilliger24thaumato} this is done in a separate flattening stage using 2D parameterization methods \cite{rabinovich17tog,levy02tog,sheffer01ewc}.
However, our spiral \textit{already} defines a 2D parameterization of the space---the canonical space $\theta$ and $z$ directions correspond to the desired U and V coordinates.
These are encouraged to align with the true orientation of the papyrus by the fiber orientation loss described above.
In practice, we can also apply SLIM~\cite{rabinovich17tog} as an extra stage to improve the smoothness of our flattening.
Given the surface mesh with UV coordinates, it is straightforward to reconstruct an unrolled volume by sampling from the original scan volume.

\section{Results}
\label{sec:results}

In this section, we thoroughly evaluate the design space of our method, considering different variants of losses, available input features, and transform parameterizations. We also compare against the only existing automated unrolling approach suitable for this data \cite{schilliger24thaumato}.
\footnote{Our code is available at \url{https://github.com/pmh47/spiral-fitting}}

\paragraph{Metrics.}
We use five metrics to quantify performance:
\begin{itemize}
    \item \textbf{WJF} (winding jump fraction) measures how often the ground-truth surfaces cross between different windings of our spiral. For 100 slices along the $z$-axis, we find line-segments where the ground-truth surface mesh intersects those slices. We then calculate the fraction of line-segments for which the nearest spiral winding at one end differs from that at the other.
    \item \textbf{MRWD} (mean radial winding distance) measures how close windings are to their ground-truth radii. It is the mean over windings, of distance from the $K$\textsuperscript{th} spiral winding to the $K$\textsuperscript{th} ground-truth winding, measured radially from the umbilicus at 100 angles per slice.
    \item \textbf{ChD} is the one-directional chamfer distance from the ground-truth surface to the predicted surface.
    \item \textbf{AD} is the mean angular defect over mesh vertices;
    this is a discrete approximation of the absolute intrinsic Gaussian curvature of the surface. It should be zero for developable surfaces (\ie those which can be flattened, which we expect to be true of the papyrus sheet).
    \item \textbf{Str} measures the amount of in-sheet stretching, given by the average over mesh edges of the difference between the edge length in scan space and UV space.
\end{itemize}

\paragraph{Quantitative results.}
We first benchmark our method against ThaumatoAnakalyptor~\cite{schilliger24thaumato}, the only existing work to unroll substantial regions of Herculaneum papyri fully automatically.
This bottom-up approach extracts surface fragments from a point-cloud, then stitches these into a surface.
Since ThaumatoAnakalyptor directly outputs a mesh, we only report ChD, AD and Str, not those metrics which rely on access to the canonical spiral representation.
Results on \phpf{} are given in Table~\ref{tab:loss-ablations}; our method (top row) outperforms \cite{schilliger24thaumato} (bottom row) according to ChD and AD, indicating closer conformance to the ground-truth surface.
Ours stretches the surface slightly more (higher Str); however if like \cite{schilliger24thaumato} we use SLIM~\cite{rabinovich17tog} flattening in addition (2nd row), this gap closes substantially.

We also measure the benefit brought by each of our losses. Rows 3--5 of Table~\ref{tab:loss-ablations} ablate respectively the surface normals, relative winding number, and distance (in the second half of training) losses; we see that the full model gives the best balance of performance across metrics.
The next four rows remove the losses defined on papyrus fibers (rather than surfaces);
removing the orientation losses has minimal effect on metrics, but causes the text to be distorted in-plane;
removing all fiber losses worsens ChD and WJF.

Next we investigate the impact of changing the diffeomorphism parameterization.
In Table~\ref{tab:transform-ablations}, we compare our full model against two ablations, removing the gap expander and per-slice scale.
We see that even though the integrated flow field can theoretically represent general diffeomorphisms, the composition with the other stages improves WJF significantly in practice.

\paragraph{Qualitative results.}
Fig.~\ref{fig:s1-sections} shows the initial (not deformed) spiral and the final deformed spiral, overlaid on several cross sections of \phpf{}; Fig.~\ref{fig:s5-results} (top) shows the same for \phost{}.
The deformed spiral windings closely conform to the distorted shape of the scroll windings in each slice.
Since the deformed spiral forms a single continuous surface, it propagates across breaks in the papyrus, bridging these regions plausibly with windings.

\setlength{\intextsep}{4pt}%
\setlength{\columnsep}{8pt}%
\begin{wrapfigure}{r}{0.48\linewidth}
    \includegraphics[width=1\linewidth]{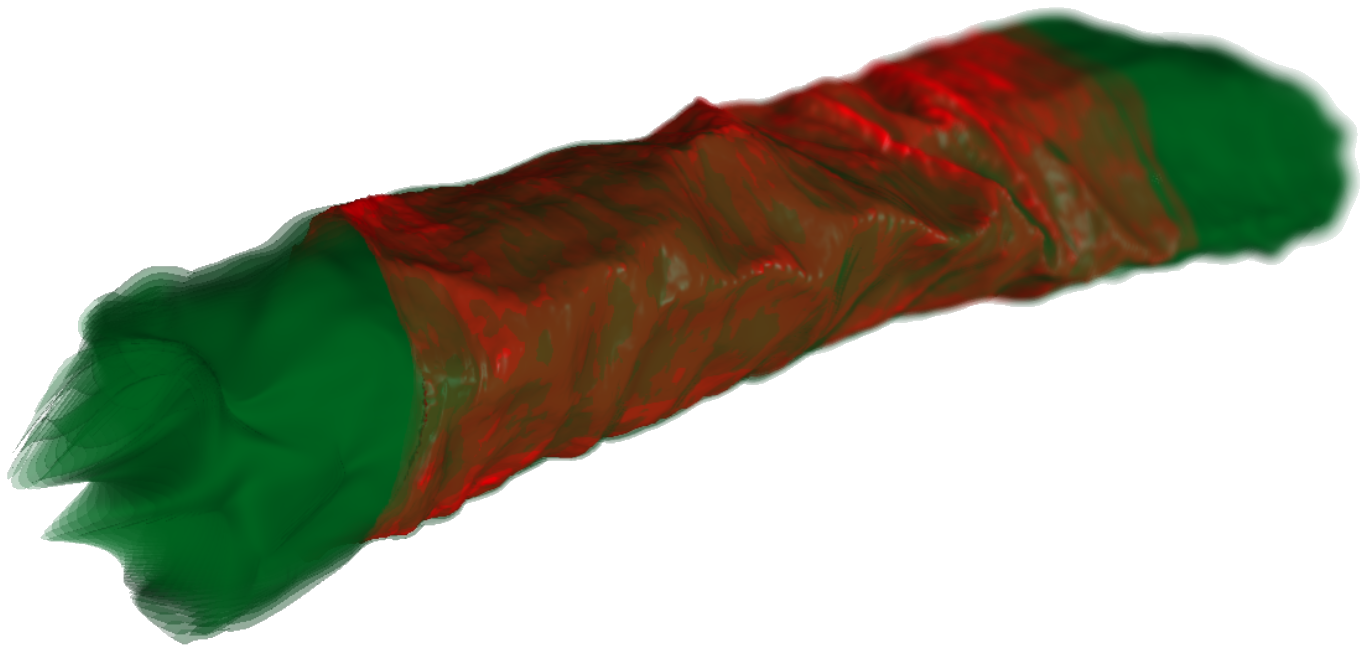}

    \vspace{-6pt}
    \small
    \caption{
    Our predicted (green) and ground-truth (red) surfaces overlaid. The surfaces are closely matched, and ours has substantially greater coverage. 
    }
    \label{fig:our-mesh-on-gp}
\end{wrapfigure}
Fig.~\ref{fig:our-mesh-on-gp} shows the ground-truth (red) and fitted (green) surfaces for \phpf{}.
Ours has significantly larger coverage than the ground-truth, and in regions where they overlap the two surfaces are closely aligned.
Fig.~\ref{fig:render-and-ink} (top) shows the unrolled surface of \phpf{}.
We can see the structure of the papyrus, with horizontal and vertical fibers visible.
Dark regions are where there is no papyrus under the predicted surface; this may be because it is broken or burned away.
Fig.~\ref{fig:render-and-ink} (bottom) shows the result of running a pretrained TimeSformer~\cite{bertasius21icml} ink detection model \cite{nader23gpw} on the unrolled scroll.
This reveals many letters forming complete words and lines of Greek text.
In the rightmost panel of Fig.~\ref{fig:s1-sections}---here instead of flattening the scroll, we apply the inverse diffeomorphism to the scan volume, to transform the scan into the canonical space.
This converts the highly distorted windings into near-concentric circles, indicating that our approach can successfully recover even the extreme distortions observed in \phpf{}.
Fig.~\ref{fig:s5-results} shows results on \phost{}; here the deformed spiral (top) covers almost the entire radius of the scroll, conforming closely with the windings.
Fig.~\ref{fig:s5-results} (bottom) shows close-ups of the scroll surface (full version in supplementary); this reveals numerous Greek letters, bright against the darker background.

\begin{table}[t]
    \centering
    \begin{tabular}{@{}ccc ccccc@{}}
    \toprule
        Flow & Aff & GE & WJF & MRWD & ChD & AD & Str \\
    \cmidrule{1-3}\cmidrule(l){4-8}
        $\checkmark$ & $\checkmark$ & $\checkmark$ & 3.20\% & 7.64 & 5.29 & 0.0568 & 1.239 \\ 
        $\checkmark$ & $\checkmark$ & ~ & 3.59\% & 6.12 & 4.59 & 0.0599 & 1.111 \\ 
        $\checkmark$ & ~ & ~ & 3.43\% & 5.84 & 4.78 & 0.0529 & 1.105 \\ 
    \bottomrule
    \end{tabular}
    \vspace{-6pt}
    \caption{Variants of our model with different parametrisations of the diffeomorphism from canonical to scroll space. \textit{Flow} is the flow field; \textit{Aff} is the z-varying affine transform; \textit{GE} is the gap-expander. Lower is better for all metrics}
    \label{tab:transform-ablations}
\end{table}

\paragraph{Computational cost.}
On a workstation with one Nvidia RTX 3090 GPU, our approach takes 19 hours for
\phpf{}.
This includes 16~hrs to postprocess the existing nnUNet outputs.
Spiral-fitting takes around 3 hours for $20\,000$ iterations, by which point the loss has converged.
The high speed of spiral fitting is due to our GPU-friendly implementation, and the design of the feature postprocessing strategy, which produces sets of 1D tracks---no volumetric data (neither raw scroll scan nor nnUNet predictions) needs to be loaded during fitting.

\section{Conclusion}
\label{sec:conclusion}

We have presented an automated approach to virtual unrolling, applied to two challenging Herculaneum Papyri.
We fit a global model of the deformed scroll to neural network predictions of where the papyrus lies, guaranteeing a single continuous 2D surface as output.
We hope our method can accelerate the unrolling of the hundreds of scrolls in the collection that have not yet been read.

\paragraph{Limitations.}
Our method minimizes a sum of L1 losses capturing how well the surface conforms to the paths; when paths are contradictory due to imperfect U-Net predictions, the surface sometimes wanders between two true windings, instead of committing to one. Using an $L^p$-norm for $p<1$, or using the augmented Lagrangian method, would therefore be fruitful directions for future work.
We also currently assume the scroll is a single rolled sheet.
It would be interesting to extend the approach to documents that are folded rather than rolled (or indeed folded then rolled), such as the Elephantine Papyri \cite{mahnke20jch}.
In principle this simply requires using a different idealized structure in the canonical space.

{
    \small
    \bibliographystyle{ieeenat_fullname}
    \bibliography{main}

@String(ECCV= {Eur. Conf. Comput. Vis.})

@String(NIPS= {Adv. Neural Inform. Process. Syst.})

@String(ICLR = {Int. Conf. Learn. Represent.})

@String(ECCV  = {ECCV})

@String(NIPS  = {NeurIPS})

@String(ICLR  = {ICLR})

@misc{parsons23dataset,
title = {{EduceLab-Scrolls: Verifiable Recovery of Text from Herculaneum Papyri using X-ray CT}},
author = {Stephen Parsons and Parker, C. Seth and Christy Chapman and Mami Hayashida and Seales, W. Brent},
year = {2023},
note = {arXiv:2304.02084},
archivePrefix = {arXiv},
primaryClass = {cs.CV},
doi = {10.48550/arXiv.2304.02084},
url = {https://arxiv.org/abs/2304.02084}
}

@software{parkerXXvolcart,
  author       = {Parker, C. Seth and
                  Gessel, Kristina and
                  Parsons, Stephen and
                  Chappell, Jacob and
                  Athie Teruel, Bruno and
                  Bentley, Nikki and
                  Bertelsman, Ali and
                  Broadbent, John and
                  Chapman, Sydney and
                  Coleman, Abigail and
                  Du, Chao and
                  Gardella, Callie and
                  Graczyk, Nick and
                  Hatch, Hannah and
                  Hogg, Lula and
                  Karlage, Sean and
                  Nguyen, Tam and
                  Pack, James and
                  Pennington, David and
                  Revers, Allison and
                  Roup, Mike and
                  Royal, Michael and
                  Seevers, Kyra and
                  Shankle, Melissa and
                  Syamil, Raiffa and
                  Taber, Ryan and
                  Posma, JP and
                  Schilliger, Julian},
  title        = {{Volume Cartographer}},
  month        = feb,
  year         = 2025,
  publisher    = {Zenodo},
  version      = {v2.27.0},
  doi          = {10.5281/zenodo.14878458},
  note          = {\url{https://doi.org/10.5281/zenodo.14878458}},
  swhid        = {swh:1:dir:29489f0eddbafd3cffc0b305945efcfe71dd05b4
                   ;origin=https://doi.org/10.5281/zenodo.4604881;vis
                   it=swh:1:snp:ab2cb7b2867f0186868ed020ba1a785129fb6
                   e5b;anchor=swh:1:rel:8927cef3c12d58e5c320d6d514408
                   02b365539c8;path=educelab-volume-
                   cartographer-9b9bfd9
                  },
}

@inproceedings{ronneberger15unet,
  author    = {Olaf Ronneberger and
               Philipp Fischer and
               Thomas Brox},
  title     = {U-Net: Convolutional Networks for Biomedical Image Segmentation},
  booktitle   = {MICCAI},
  year      = {2015}, 
}

@article{isensee21nnunet,
  author = {Isensee, Fabian and Jaeger, Paul F. and Kohl, Simon A. A. and Petersen, Jens and Maier-Hein, Klaus H.},
  title = {{nnU-Net}: a self-configuring method for deep learning-based biomedical image segmentation},
  journal = {Nature Methods},
  year = {2021},
  month = {2},
  volume = {18},
  number = {2},
  pages = {203--211},
  doi = {10.1038/s41592-020-01008-z},
  url = {https://doi.org/10.1038/s41592-020-01008-z},
  issn = {1548-7105}
}

@INPROCEEDINGS {klenert25wacvw,
author = { Klenert, Nicolas and Schwoerer, Finn and Hajarolasvadi, Noushin and Bournez, Siloe and Arlt, Tobias and Mahnke, Heinz-Eberhard and Lepper, Verena and Baum, Daniel },
booktitle = { 2025 IEEE/CVF Winter Conference on Applications of Computer Vision Workshops (WACVW) },
title = {{ Improving the Identification of Layers in 3D Images of Ancient Papyrus Using Artificial Neural Networks }},
year = {2025},
pages = {1204-1212},
doi = {10.1109/WACVW65960.2025.00143},
url = {https://doi.ieeecomputersociety.org/10.1109/WACVW65960.2025.00143},
publisher = {IEEE Computer Society},
}

@misc{silversmith21cc3d,
author = {Silversmith, William},
note = {\url{https://zenodo.org/record/5535251}},
month = sep,
title = {{cc3d: Connected components on multilabel 3D \& 2D images.}},
version = {3.2.1},
year = {2021}
}

@article{rosenfeld66cc,
author = {Rosenfeld, Azriel and Pfaltz, John L.},
title = {Sequential Operations in Digital Picture Processing},
year = {1966},
issue_date = {Oct. 1966},
publisher = {Association for Computing Machinery},
address = {New York, NY, USA},
volume = {13},
number = {4},
issn = {0004-5411},
url = {https://doi.org/10.1145/321356.321357},
doi = {10.1145/321356.321357},
journal = {J. ACM},
month = oct,
pages = {471–494},
numpages = {24}
}

@techreport{wu05cc,
author={K. Wu and E. Otoo and K. Suzuki},
title={Two Strategies to Speed up Connected Component Labeling Algorithms},
year={2005},
institution={Lawrence Berkeley National Laboratory},
number={LBNL-29102}
}

@software{silversmith21kimi,
  author = {Silversmith, William and Bae, J. Alexander and Li, Peter H. and Wilson, A. M.},
  title = {{Kimimaro: Skeletonize densely labeled 3D image segmentations}},
  version = {3.0.0},
  date = {2021-09-29},
  doi = {10.5281/zenodo.5539913},
  note = {\url{https://doi.org/10.5281/zenodo.5539913}}
}

@INPROCEEDINGS{sato00teasar,
  author={Sato, M. and Bitter, I. and Bender, M.A. and Kaufman, A.E. and Nakajima, M.},
  booktitle={Proceedings the Eighth Pacific Conference on Computer Graphics and Applications}, 
  title={{TEASAR}: tree-structure extraction algorithm for accurate and robust skeletons}, 
  year={2000},
  pages={281-449},
  doi={10.1109/PCCGA.2000.883951}
}

@inproceedings{bertasius21icml,
    author  = {Gedas Bertasius and Heng Wang and Lorenzo Torresani},
    title = {Is Space-Time Attention All You Need for Video Understanding?},
    booktitle   = {Proceedings of the International Conference on Machine Learning (ICML)}, 
    month = {July},
    year = {2021}
}

@misc{nader23gpw,
  author = {Youssef Nader and Luke Farritor and Julian Schilliger},
  title = {{Vesuvius-Grandprize-Winner}},
  version = {0efba1b5},
  date = {2024-11-25},
  note = {\url{https://github.com/younader/Vesuvius-Grandprize-Winner}}
}

@misc{schilliger24thaumato,
  author = {Julian Schilliger},
  title = {{ThaumatoAnakalyptor}},
  date = {2024-11},
  note = {\url{https://github.com/schillij95/ThaumatoAnakalyptor}}
}

@book{sider05book,
  title        = {The Library of the Villa dei Papiri at Herculaneum},
  author       = {David Sider},
  year         = {2005},
  publisher    = {J. Paul Getty Museum},
  address      = {Los Angeles},
  isbn         = {978-0-89236-799-3},
}

@book{barker1908book,
  author       = {Barker, Ethel Ross},
  title        = {Buried Herculaneum},
  year         = {1908},
  publisher    = {A. and C. Black},
  address      = {London},
}

@article{bics86,
title = {{IV.~The Herculaneum Papyri}},
journal = {Bulletin of the Institute of Classical Studies},
volume = {33},
number = {S54},
pages = {36-45},
doi = {https://doi.org/10.1111/j.2041-5370.1986.tb01374.x},
url = {https://onlinelibrary.wiley.com/doi/abs/10.1111/j.2041-5370.1986.tb01374.x},
year = {1986}
}

@book{lambert97traces,
title={Traces of the past: unraveling the secrets of archaeology through chemistry},
author={Lambert, Joseph B},
year={1997},
publisher={Perseus Books},
address = {Reading, Mass.}
}

@article{wallert89papyrus,
author = {A. Wallert},
title = {The reconstruction of papyrus manufacture: a preliminary investigation},
journal = {Studies in Conservation},
volume = {34},
number = {1},
pages = {1--8},
year = {1989},
publisher = {Routledge},
doi = {10.1179/sic.1989.34.1.1},
}

@inproceedings{seales04jcdl,
author = {Seales, W. B. and Lin, Yun},
title = {Digital restoration using volumetric scanning},
year = {2004},
isbn = {1581138326},
publisher = {Association for Computing Machinery},
address = {New York, NY, USA},
url = {https://doi.org/10.1145/996350.996380},
doi = {10.1145/996350.996380},
booktitle = {Proceedings of the 4th ACM/IEEE-CS Joint Conference on Digital Libraries},
pages = {117–124},
numpages = {8},
location = {Tuscon, AZ, USA},
series = {JCDL '04}
}

@article{seales16sciadv,
author = {William Brent Seales  and Clifford Seth Parker  and Michael Segal  and Emanuel Tov  and Pnina Shor  and Yosef Porath },
title = {From damage to discovery via virtual unwrapping: Reading the scroll from {En-Gedi}},
journal = {Science Advances},
volume = {2},
number = {9},
pages = {e1601247},
year = {2016},
doi = {10.1126/sciadv.1601247},
URL = {https://www.science.org/doi/abs/10.1126/sciadv.1601247},
}

@article{nicolardi24zfpe,
  author       = {Federica Nicolardi and Stephen Parsons and Daniel Delattre and Gianluca Del Mastro and Robert L. Fowler and Richard Janko and Tobias Reinhardt and C.~Seth Parker and Christy Chapman and W.~Brent Seales},
  title        = {{Revealing Text from a Still‑Rolled Herculaneum Papyrus Scroll (PHerc.~Paris.~4)}},
  journal      = {Zeitschrift für Papyrologie und Epigraphik},
  number       = {229},
  pages        = {1--13},
  month        = mar,
  year         = {2024},
  publisher    = {Dr. Rudolf Habelt GmbH},
  issn         = {0084‑5388},
}

@inproceedings{parsons2020icch,
  author       = {Stephen Parsons and Kyle Gessel and C. Seth Parker and W. Brent Seales},
  title        = {Deep Learning for More Expressive Virtual Unwrapping},
  booktitle    = {Proceedings of the 25th International Conference on Cultural Heritage and New Technologies},
  address      = {Heidelberg},
  year         = {2020},
  pages        = {203--207},
}

@article{dambrogio21natcom,
  author       = {Dambrogio, Jana and Ghassaei, Amanda and Smith, Daniel Starza and Jackson, Holly and Demaine, Martin L. and Davis, Graham and Mills, David and Ahrendt, Rebekah and Akkerman, Nadine and van der Linden, David and Demaine, Erik D.},
  title        = {Unlocking history through automated virtual unfolding of sealed documents imaged by X-ray microtomography},
  journal      = {Nature Communications},
  volume       = {12},
  number       = {1184},
  year         = {2021},
  month        = mar,
  doi          = {10.1038/s41467-021-21326-w},
  issn         = {2041-1723},
  license      = {CC BY 4.0}
}

@article{mocella15natcomm,
  author = {Mocella, Vito and Brun, Emmanuel and Ferrero, Claudio and Delattre, Daniel},
  title = {{Revealing letters in rolled Herculaneum papyri by X-ray phase-contrast imaging}},
  journal = {Nature Communications},
  year = {2015},
  month = jan,
  volume = {6},
  number = {1},
  pages = {5895},
  issn = {2041-1723},
  doi = {10.1038/ncomms6895},
  url = {https://doi.org/10.1038/ncomms6895}
}

@article{seales13virtual,
  author = {Seales, W. B. and Delattre, D.},
  journal = {Cronache Ercolanesi},
  pages = {191-208},
  title = { Virtual unrolling of carbonized {Herculaneum scrolls}: Research Status (2007–2012) },
  volume = 43,
  year = 2013
}

@article{stabile21scirep,
  author = {Stabile, Sara and Palermo, Francesca and Bukreeva, Inna and Mele, Daniela and Formoso, Vincenzo and Bartolino, Roberto and Cedola, Alessia},
  title = {A computational platform for the virtual unfolding of {Herculaneum Papyri}},
  journal = {Scientific Reports},
  year = {2021},
  month = {1},
  day = {18},
  volume = {11},
  number = {1695},
  issn = {2045-2322},
  doi = {10.1038/s41598-020-80458-z},
  url = {https://doi.org/10.1038/s41598-020-80458-z},
}

@incollection{capasso20philodemus,
    author = {Capasso, Mario},
    isbn = {9780199744213},
    title = {{Philodemus and the Herculaneum Papyri}},
    booktitle = {Oxford Handbook of Epicurus and Epicureanism},
    publisher = {Oxford University Press},
    year = {2020},
    month = {08},
    pages = {378--429},
    doi = {10.1093/oxfordhb/9780199744213.013.13},
    url = {https://doi.org/10.1093/oxfordhb/9780199744213.013.13},
}

@article{parker19plosone,
    doi = {10.1371/journal.pone.0215775},
    author = {Parker, Clifford Seth AND Parsons, Stephen AND Bandy, Jack AND Chapman, Christy AND Coppens, Frederik AND Seales, William Brent},
    journal = {PLOS ONE},
    publisher = {Public Library of Science},
    title = {From invisibility to readability: Recovering the ink of {Herculaneum}},
    year = {2019},
    month = {05},
    volume = {14},
    url = {https://doi.org/10.1371/journal.pone.0215775},
    pages = {1-17},
    number = {5},
}

@article{levy02tog,
author = {Levy, Bruno and Petitjean, Sylvain and Ray, Nicolas and Maillot, J\'{e}rome},
title = {Least squares conformal maps for automatic texture atlas generation},
year = {2002},
publisher = {Association for Computing Machinery},
address = {New York, NY, USA},
booktitle = {ACM Transaction on Graphics},
volume={21},
number={3},
pages={362--371}
}

@article{sheffer01ewc,
  author = {Sheffer, A. and de Sturler, E.},
  year = {2001},
  month = {oct},
  title = {Parameterization of Faceted Surfaces for Meshing using Angle-Based Flattening},
  journal = {Engineering with Computers},
  volume = {17},
  number = {3},
  pages = {326--337},
  issn = {1435-5663},
  doi = {10.1007/PL00013391},
  url = {https://doi.org/10.1007/PL00013391}
}

@ARTICLE{klenert24tvcg,
  author={Klenert, Nicolas and Lepper, Verena and Baum, Daniel},
  journal={IEEE Transactions on Visualization and Computer Graphics}, 
  title={A Local Iterative Approach for the Extraction of 2D Manifolds from Strongly Curved and Folded Thin-Layer Structures}, 
  year={2024},
  volume={30},
  number={1},
  pages={1260-1270},
  doi={10.1109/TVCG.2023.3327403}
}

@article{ashburner07ni,
title = {A fast diffeomorphic image registration algorithm},
journal = {NeuroImage},
volume = {38},
number = {1},
pages = {95-113},
year = {2007},
issn = {1053-8119},
doi = {https://doi.org/10.1016/j.neuroimage.2007.07.007},
url = {https://www.sciencedirect.com/science/article/pii/S1053811907005848},
author = {John Ashburner},
}

@article{vercauteren09ni,
title = {Diffeomorphic demons: Efficient non-parametric image registration},
journal = {NeuroImage},
volume = {45},
number = {1, Supplement 1},
pages = {S61-S72},
year = {2009},
note = {Mathematics in Brain Imaging},
issn = {1053-8119},
doi = {https://doi.org/10.1016/j.neuroimage.2008.10.040},
url = {https://www.sciencedirect.com/science/article/pii/S1053811908011683},
author = {Tom Vercauteren and Xavier Pennec and Aymeric Perchant and Nicholas Ayache},
}

@article{moler03siam,
author = {Moler, Cleve and Van Loan, Charles},
title = {Nineteen Dubious Ways to Compute the Exponential of a Matrix, Twenty-Five Years Later},
journal = {SIAM Review},
volume = {45},
number = {1},
pages = {3-49},
year = {2003},
doi = {10.1137/S00361445024180},
URL = {https://doi.org/10.1137/S00361445024180},
eprint = {https://doi.org/10.1137/S00361445024180 }
}

@article{christensen94,
doi = {10.1088/0031-9155/39/3/022},
url = {https://dx.doi.org/10.1088/0031-9155/39/3/022},
year = {1994},
month = {mar},
publisher = {},
volume = {39},
number = {3},
pages = {609},
author = {Christensen, G E and Rabbitt R D and Miller M I},
title = {3D brain mapping using a deformable neuroanatomy},
journal = {Physics in Medicine \& Biology},
}

@InProceedings{arsigny06,
author="Arsigny, Vincent
and Commowick, Olivier
and Pennec, Xavier
and Ayache, Nicholas",
editor="Pluim, Josien P. W.
and Likar, Bo{\v{s}}tjan
and Gerritsen, Frans A.",
title="A Log-Euclidean Polyaffine Framework for Locally Rigid or Affine Registration",
booktitle="Biomedical Image Registration",
year="2006",
publisher="Springer Berlin Heidelberg",
address="Berlin, Heidelberg",
pages="120--127",
isbn="978-3-540-35649-3"
}

@article{dalca19mia,
title = {Unsupervised learning of probabilistic diffeomorphic registration for images and surfaces},
journal = {Medical Image Analysis},
volume = {57},
pages = {226-236},
year = {2019},
issn = {1361-8415},
doi = {https://doi.org/10.1016/j.media.2019.07.006},
url = {https://www.sciencedirect.com/science/article/pii/S1361841519300635},
author = {Dalca, Adrian V. and Guha Balakrishnan and John Guttag and Sabuncu, Mert R.},
}

@inproceedings{kingma16neurips,
    author = {Kingma, Diederik P. and Tim Salimans and Rafal Jozefowicz and Xi Chen and Ilya Sutskever and Max Welling},
    title = {Improving Variational Inference with Inverse Autoregressive Flow},
    booktitle = {NIPS},
    year = {2016}
}

@inproceedings{rezende15icml,
    author = {Rezende, Danilo Jimenez and Shakir Mohamed},
    title = {Variational Inference with Normalizing Flows},
    booktitle = {ICML},
    year = {2015}
}

@article{papamakarios21jmlr,
  author  = {George Papamakarios and Eric Nalisnick and Rezende, Danilo Jimenez and Shakir Mohamed and Balaji Lakshminarayanan},
  title   = {Normalizing Flows for Probabilistic Modeling and Inference},
  journal = {Journal of Machine Learning Research},
  year    = {2021},
  volume  = {22},
  number  = {57},
  pages   = {1--64},
  url     = {http://jmlr.org/papers/v22/19-1028.html}
}

@inproceedings{chen18neurips,
  title={Neural Ordinary Differential Equations},
  author={Chen, Ricky T. Q. and Rubanova, Yulia and Bettencourt, Jesse and Duvenaud, David K},
  booktitle={Advances in Neural Information Processing Systems 32},
  year={2018}
}

@inproceedings{grathwohl18iclr,
  title={{FFJORD}: Free-form continuous dynamics for scalable reversible generative models},
  author={Grathwohl, Will and Chen, Ricky T. Q. and Bettencourt, Jesse and Sutskever, Ilya and Duvenaud, David},
  booktitle={International Conference on Learning Representations},
  year={2019}
}

@article{bukreeva16scirep,
  author = {Bukreeva, I. and Mittone, A. and Bravin, A. and Festa, G. and Alessandrelli, M. and Coan, P. and Formoso, V. and Agostino, R. G. and Giocondo, M. and Ciuchi, F. and Fratini, M. and Massimi, L. and Lamarra, A. and Andreani, C. and Bartolino, R. and Gigli, G. and Ranocchia, G. and Cedola, A.},
  title = {Virtual unrolling and deciphering of {Herculaneum} papyri by X-ray phase-contrast tomography},
  journal = {Scientific Reports},
  year = {2016},
  month = {jun},
  volume = {6},
  number = {1},
  pages = {27227},
  issn = {2045-2322},
  doi = {10.1038/srep27227},
  url = {https://doi.org/10.1038/srep27227}
}

@InProceedings{kulagin24icdar,
author="Kulagin, Petr
and Polevoy, Dmitry
and Chukalina, Marina
and Nikolaev, Dmitry
and Arlazarov, Vladimir V.",
editor="Barney Smith, Elisa H.
and Liwicki, Marcus
and Peng, Liangrui",
title="Fully Automatic Virtual Unwrapping Method for Documents Imaged by X-Ray Tomography",
booktitle="Document Analysis and Recognition - ICDAR 2024",
year="2024",
publisher="Springer Nature Switzerland",
address="Cham",
pages="233--250",
isbn="978-3-031-70543-4"
}

@misc{polevoy22ctocr,
  author       = {Polevoy, Dmitry and
                  Kulagin, Petr and
                  Ingacheva, Anastasia and
                  Soldatova, Zhanna and
                  Chukalina, Marina and
                  Nikolaev, Dmitry and
                  Arlazarov, Vladimir V.},
  title        = {{CT-OCR-2022}},
  month        = oct,
  year         = 2022,
  publisher    = {Zenodo},
  doi          = {10.5281/zenodo.7324269},
  note          = {\url{https://doi.org/10.5281/zenodo.7324269}},
}

@inproceedings{schult23icra,
  title     = {{Mask3D: Mask Transformer for 3D Semantic Instance Segmentation}},
  author    = {Schult, Jonas and Engelmann, Francis and Hermans, Alexander and Litany, Or and Tang, Siyu and Leibe, Bastian},
  booktitle = {{International Conference on Robotics and Automation (ICRA)}},
  year      = {2023}
}

@ARTICLE{steger98pami,
  author={Steger, C.},
  journal={IEEE Transactions on Pattern Analysis and Machine Intelligence}, 
  title={An unbiased detector of curvilinear structures}, 
  year={1998},
  volume={20},
  number={2},
  pages={113-125},
  doi={10.1109/34.659930}
}

@article{baum21jerash,
title = {Revisiting the {Jerash Silver Scroll}: A new visual data analysis approach},
journal = {Digital Applications in Archaeology and Cultural Heritage},
volume = {21},
pages = {e00186},
year = {2021},
issn = {2212-0548},
doi = {https://doi.org/10.1016/j.daach.2021.e00186},
url = {https://www.sciencedirect.com/science/article/pii/S2212054821000151},
author = {Daniel Baum and Felix Herter and John Møller Larsen and Achim Lichtenberger and Rubina Raja},
}

@inproceedings{allegra15eusipco,
  title={Virtual unrolling using x-ray computed tomography},
  author={Allegra, Dario and Ciliberto, Enrico and Ciliberto, Paolo and Milotta, Filippo Luigi Maria and Petrillo, Giuseppe and Stanco, Filippo and Trombatore, C},
  booktitle={23rd European Signal Processing Conference (EUSIPCO)},
  pages={2864--2868},
  year={2015},
  organization={IEEE}
}

@article{mahnke20jch,
title = {Virtual unfolding of folded papyri},
journal = {Journal of Cultural Heritage},
volume = {41},
pages = {264-269},
year = {2020},
issn = {1296-2074},
doi = {https://doi.org/10.1016/j.culher.2019.07.007},
url = {https://www.sciencedirect.com/science/article/pii/S1296207419301670},
author = {Heinz-Eberhard Mahnke and Tobias Arlt and Daniel Baum and Hans-Christian Hege and Felix Herter and Norbert Lindow and Ingo Manke and Tzulia Siopi and Eve Menei and Marc Etienne and Verena Lepper},
}

@article{baum17apa,
  author = {Baum, Daniel and Lindow, Norbert and Hege, Hans-Christian and Lepper, Verena and Siopi, Tzulia and Kutz, Frank and Mahlow, Kristin and Mahnke, Heinz-Eberhard},
  title = {Revealing hidden text in rolled and folded papyri},
  journal = {Applied Physics A},
  year = {2017},
  month = {feb},
  volume = {123},
  number = {3},
  pages = {171},
  issn = {1432-0630},
  doi = {10.1007/s00339-017-0808-6},
  url = {https://doi.org/10.1007/s00339-017-0808-6}
}

@article{hoffmannbarfod15scirep,
  author   = {Hoffmann Barfod, Gry and Larsen, John Møller and Lichtenberger, Achim and Raja, Rubina},
  title    = {Revealing text in a complexly rolled silver scroll from Jerash with computed tomography and advanced imaging software},
  journal  = {Scientific Reports},
  year     = {2015},
  volume   = {5},
  number   = {17765},
  month    = {dec},
  doi      = {10.1038/srep17765},
  url      = {https://doi.org/10.1038/srep17765},
  issn     = {2045-2322},
}

@ARTICLE{schultz09tvcg,
  author={Schultz, Thomas and Theisel, Holger and Seidel, Hans-Peter},
  journal={IEEE Transactions on Visualization and Computer Graphics}, 
  title={Crease Surfaces: From Theory to Extraction and Application to Diffusion Tensor {MRI}}, 
  year={2010},
  volume={16},
  number={1},
  pages={109-119},
  doi={10.1109/TVCG.2009.44}
}

@ARTICLE{algarni19pami,
  author={Algarni, Marei and Sundaramoorthi, Ganesh},
  journal={IEEE Transactions on Pattern Analysis and Machine Intelligence}, 
  title={{SurfCut}: Surfaces of Minimal Paths from Topological Structures}, 
  year={2019},
  volume={41},
  number={3},
  pages={726-739},
  doi={10.1109/TPAMI.2018.2811810}
}

@ARTICLE{fouard06tmi,
	author = {Fouard, Céline and Malandain, Grégoire and Prohaska, Steffen and Westerhoff, Malte},
	title = {Blockwise processing applied to brain microvascular network study},
	year = {2006},
	journal = {IEEE Transactions on Medical Imaging},
	volume = {25},
	number = {10},
	pages = {1319 – 1328},
	doi = {10.1109/TMI.2006.880670},
}

@article{otsu79,
  author    = {Otsu, Nobuyuki},
  title     = {A Threshold Selection Method from Gray-Level Histograms},
  journal   = {IEEE Transactions on Systems, Man, and Cybernetics},
  year      = {1979},
  volume    = {9},
  number    = {1},
  pages     = {62--66},
  doi       = {10.1109/TSMC.1979.4310076},
  publisher = {IEEE}
}

@article{eberly94jmi,
  author    = {Eberly, D. and Gardner, R. and Morse, B. and Pizer, S. and Scharlach, C.},
  title     = {Ridges for image analysis},
  journal   = {Journal of Mathematical Imaging and Vision},
  year      = {1994},
  month     = {dec},
  volume    = {4},
  pages     = {353--373},
  doi       = {10.1007/BF01262402},
  url       = {https://doi.org/10.1007/BF01262402},
  issn      = {1573-7683},
}

@misc{schilling25fasp,
  author = {Hendrik Schilling},
  title = {{FASP}},
  date = {2025-01-19},
  note = {\url{https://github.com/hendrikschilling/FASP}}
}

@inproceedings{kingma15adam,
  title={Adam: A Method for Stochastic Optimization},
  author={Kingma, Diederik P. and Ba, Jimmy},
  booktitle={Proceedings of the 3rd International Conference on Learning Representations (ICLR)},
  year={2015}
}

@article{rabinovich17tog,
author = {Rabinovich, Michael and Poranne, Roi and Panozzo, Daniele and Sorkine-Hornung, Olga},
title = {Scalable Locally Injective Mappings},
year = {2017},
issue_date = {April 2017},
publisher = {Association for Computing Machinery},
address = {New York, NY, USA},
volume = {36},
number = {2},
issn = {0730-0301},
url = {https://doi.org/10.1145/2983621},
doi = {10.1145/2983621},
journal = {ACM Trans. Graph.},
month = apr,
articleno = {16},
numpages = {16},
}

@article{wang22ole,
title = {Virtual unrolling technology based on terahertz computed tomography},
journal = {Optics and Lasers in Engineering},
volume = {151},
pages = {106924},
year = {2022},
issn = {0143-8166},
doi = {https://doi.org/10.1016/j.optlaseng.2021.106924},
url = {https://www.sciencedirect.com/science/article/pii/S0143816621003936},
author = {Tianyi Wang and Kejia Wang and Kaigang Zou and Sishi Shen and Yongqiang Yang and Mengting Zhang and Zhengang Yang and Jinsong Liu}
}

@ARTICLE{liu18tip,
  author={Liu, Chang and Rosin, Paul L. and Lai, Yu-Kun and Hu, Weiduo},
  journal={IEEE Transactions on Image Processing}, 
  title={Robust Virtual Unrolling of Historical Parchment XMT Images}, 
  year={2018},
  volume={27},
  number={4},
  pages={1914-1926},
  doi={10.1109/TIP.2017.2783626}
}

@article{samko14pr,
title = {Virtual unrolling and information recovery from scanned scrolled historical documents},
journal = {Pattern Recognition},
volume = {47},
number = {1},
pages = {248-259},
year = {2014},
issn = {0031-3203},
doi = {https://doi.org/10.1016/j.patcog.2013.06.015},
url = {https://www.sciencedirect.com/science/article/pii/S0031320313002677},
author = {Oksana Samko and Yu-Kun Lai and David Marshall and Paul L. Rosin}
}

@inproceedings{samko11bmvc,
title={Segmentation of Parchment Scrolls
for Virtual Unrolling},
author={Oksana Samko and Yu-Kun Lai and David Marshall and Paul L. Rosin},
booktitle={British Machine Vision Conference},
year={2011}
}

@article{xie22fields,
author = {Xie, Yiheng and Takikawa, Towaki and Saito, Shunsuke and Litany, Or and Yan, Shiqin and Khan, Numair and Tombari, Federico and Tompkin, James and Sitzmann, Vincent and Sridhar, Srinath},
title = {Neural Fields in Visual Computing and Beyond},
journal = {Computer Graphics Forum},
volume = {41},
number = {2},
pages = {641-676},
doi = {https://doi.org/10.1111/cgf.14505},
url = {https://onlinelibrary.wiley.com/doi/abs/10.1111/cgf.14505},
year = {2022}
}

@inproceedings{mildenhall2020nerf,
 title={NeRF: Representing Scenes as Neural Radiance Fields for View Synthesis},
 author={Ben Mildenhall and Pratul P. Srinivasan and Matthew Tancik and Jonathan T. Barron and Ravi Ramamoorthi and Ren Ng},
 year={2020},
 booktitle={ECCV},
}
}

\section*{Appendices}

\appendix

\section{Transform parameterization}

In this section we give additional details on the three components that make up the overall diffeomorphism $T_{S \rightarrow V}$ that maps the idealized spiral in canonical space to the highly deformed shape observed in the scan volume.
The transform is a composition $T_{S\rightarrow V} = T_\mathrm{aff} \,\circ\, T_\mathrm{flow} \,\circ\, T_\mathrm{gap}$, where:
(i) $T_\mathrm{aff}$ is a per-slice affine (scaling and translation) transform;
(ii) $T_\mathrm{flow}$ is the result of integrating a flow field;
(iii) $T_\mathrm{gap}$ rescales inter-winding gaps.

\subsection{Per-slice affine transform}

We non-isotropically scale and translate each slice in the $xy$-plane; the scale factors and translations vary wrt $z$.
This stage is parameterized by $N$ 2D log-scales and translations where $N$ is a fixed number of keypoints along the $z$-axis; the translations and log-scales are linearly interpolated for other $z$ values.
Thus, for a given point $\mathbf{x}=(x_1,\,x_2,\,x_3)$, we have \begin{equation}
    T_\mathrm{aff}(\mathbf{x}) =
    \left(\begin{array}{c}
         x_1 \exp(s_1) + t_1 \\
         x_2 \exp(s_2) + t_2 \\
         x_3
    \end{array}
    \right)
\end{equation}
where $(s_1, s_2) = (1-\alpha)\, \mathbf{s}^{(\lfloor \tau \rfloor)} + \alpha\, \mathbf{s}^{(\lfloor t \rfloor+1)}, 
\quad \alpha = \tau - \lfloor \tau \rfloor$, $\tau = N (x_3 - z_\mathrm{min}) / (z_\mathrm{max} - z_\mathrm{min}$), and $\mathbf{S}^{(n)}, n = 1 \ldots N$ are the $N$ log-scales; translations are defined analogously.

\subsection{Integrated flow field}

We introduce a flow velocity field $\mathbf{u}$, \ie a 3D vector field of 3D velocities defined over the scan volume.
This allows us to transform a given point by simulating its movement in flow of the given local velocities for a fixed length of time.
This is done by solving an ODE defined by $\mathbf{u}$ and an initial condition (the starting point).
Writing $\phi^{(t)}$ for the transform induced by simulating the flow $\mathbf{u}$ for time $t$,
the final transformed position of a point $\mathbf{x}$ is given by $T_\mathrm{flow}(\mathbf{x}) = \phi^{(1)}(\mathbf{x})$, where $\phi^{(t)}$ evolves according to
\begin{equation}
    \frac{d}{dt}\phi^{(t)}(\mathbf{x}) =
    \mathbf{u}\left(
        \phi^{(t)}(\mathbf{x})
    \right)
\end{equation}
with the initial condition $\phi^{(0)}(\mathbf{x}) = \mathbf{x}$, \ie we start with an identity transform.
The resulting transformation is guaranteed to be a diffeomorphism, provided the flow field $\mathbf{u}$ is itself smooth \cite{ashburner07ni,arsigny06}.
Formally, the space of flow fields is the Lie algebra that generates the Lie group of diffeomorphisms; the former provides a straightforward way to parameterize the latter.
Moreover, the inverse diffeomorphism is easily found by negating the flow field, then integrating backwards in time.
We define the velocity field as the sum two underlying fields $\mathbf{u}=\mathbf{u}_\mathrm{coarse}+\mathbf{u}_\mathrm{fine}$, where $\mathbf{u}_\mathrm{coarse}$ is more spatially smooth than $\mathbf{u}_\mathrm{fine}$.
Both fields are defined by trilinear interpolation of explicitly-represented discrete grids of 3D velocity vectors.
The grid for $\mathbf{u}_\mathrm{fine}$ is $48\times$ lower resolution than the original CT volume, and that for $\mathbf{u}_\mathrm{coarse}$ is a further $6\times$ lower.

We solve the ODE using explicit Euler integration with 16 steps; more steps did not result in significantly different results, whereas fewer could cause $\TSV \circ \TVS$ to not be sufficiently close to the identity transform.
Experiments with scale-and-square integration \cite{moler03siam}---which requires fewer operations---showed that it is less stable in our setting than Euler integration.
To update the flow field via gradient descent, we must differentiate back through the ODE solve, i.e.~compute the Jacobian of $\phi^{(1)}(\mathbf{x})$ wrt $\mathbf{u}$.
We achieve this by directly unrolling and back-propagating through the Euler updates; this is tractable due to the relatively small number of steps. With more steps or larger minibatches, it would become preferable to use the adjoint method, similar to continuous normalizing flows \cite{chen18neurips}.

We also conducted initial experiments using a time-varying velocity field, defined explicitly by changing the 3D coarse and fine grids to be 4D. This has only a minor effect on compute cost but increases memory linearly with the number of timesteps. Despite such fields having intuitively simpler behavior, we did not notice any significant improvements to final results, hence all our main experiments use a time-constant field.

We also tested an implicit velocity field defined by a small neural network (a residual MLP) operating on 3D coordinates in the volume, and regressing these directly to the velocity vector, similar to neural radiance fields (NeRF) \cite{mildenhall2020nerf} and other neural field methods \cite{xie22fields}. While this potentially allows allocating parameter capacity more efficiently where needed (instead of uniformly across the volume regardless of curvature), in practice we found the computational expense to be prohibitive.

\subsection{Inter-winding gap scaling}

We define a scalar field over the 2D surface of the canonical scroll spiral, which specifies for each point, by how much the next winding should be pushed outward from its default distance (defined by being an Archimedean spiral), \ie locally scaling the gaps between windings.
Since this field is defined only on the 2D spiral surface (truncated at a certain number of windings), it has quadratic cost in memory. This allows using a finer resolution than the full-volume flow field, and is also more expressive since it can directly separate or push apart nearby sheets regardless of how their position relates to the fixed flow-field grids.
The field itself is defined explicitly by scalar values on a 2D grid wrt the $z$ and $\theta$ coordinates of eq.~1 in the main paper. This discrete representation is extended to the 2D plane by bilinear interpolation.
The field values are interpreted as log scale factors multiplying the spacing between windings, originally defined by the global winding rate $\omega$.
Specifically, for a given point $\mathbf{x}$, we find the next spiral winding inwards from it, measure the radial distance to that winding, multiply this distance by the relevant scale factor, and move the point outwards correspondingly.

\section{Hyperparameters}

\noindent
\begin{tabular}{@{}ll@{}}
\textbf{Diffeomorphism parameterization} \\
number of Euler integration steps & 16 \\
voxel resolution of flow field & 12 \\
voxel resolution of gap expander & 32 \\
\\
\textbf{Optimization} \\
learning rate & $5 \times 10^{-4}$ \\
learning rate schedule & constant \\
number of optimization steps & 20000 \\
num.~points sampled per path & 100 \\
num.~points for relative winding number loss & 2000 \\
num.~points for normals loss & 2000 \\
num.~points for regularization & 1500 \\
path distance loss starting from step & 10000 \\
\\
\textbf{Loss weights} \\
surface normal & 200 \\
path radius & 5 \\
relative winding number & 10 \\
path distance (L1) & 4 \\
fiber direction & 5 \\
stretch regularization & 200 \\
centerline fixed & 1 \\
\end{tabular}

\section{Additional Visualizations}

\begin{figure*}
\begin{subfigure}[b]{0.78\linewidth}
    \centering
    \includegraphics[width=\linewidth,trim={0 0 67em 0},clip]{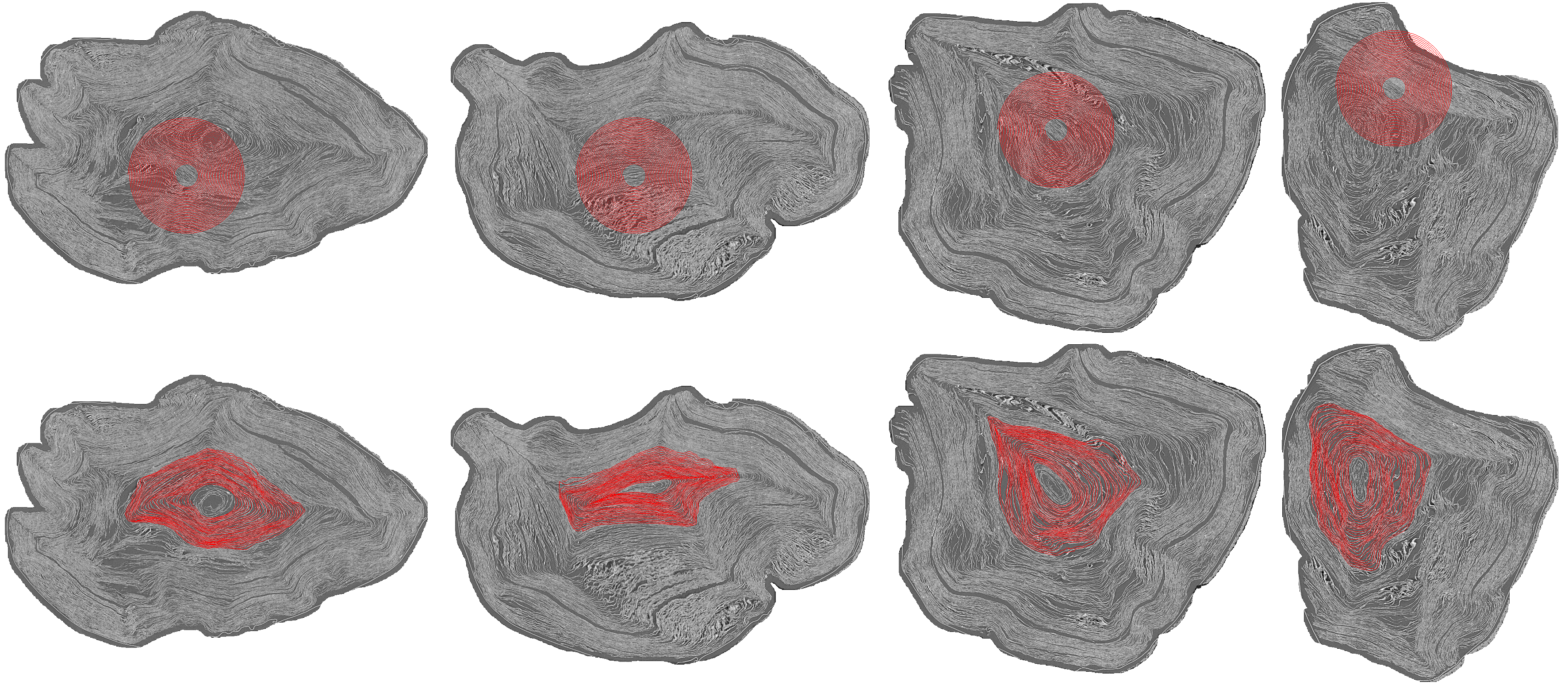}

    \vspace{-4pt}
    \caption{}
    \label{fig:s1-spiral-initial-final}
\end{subfigure}%
\hfill
\begin{subfigure}[b]{0.2\linewidth}
    \includegraphics[width=\linewidth]{figures/undeformed-slice-1.png}
    \includegraphics[width=\linewidth]{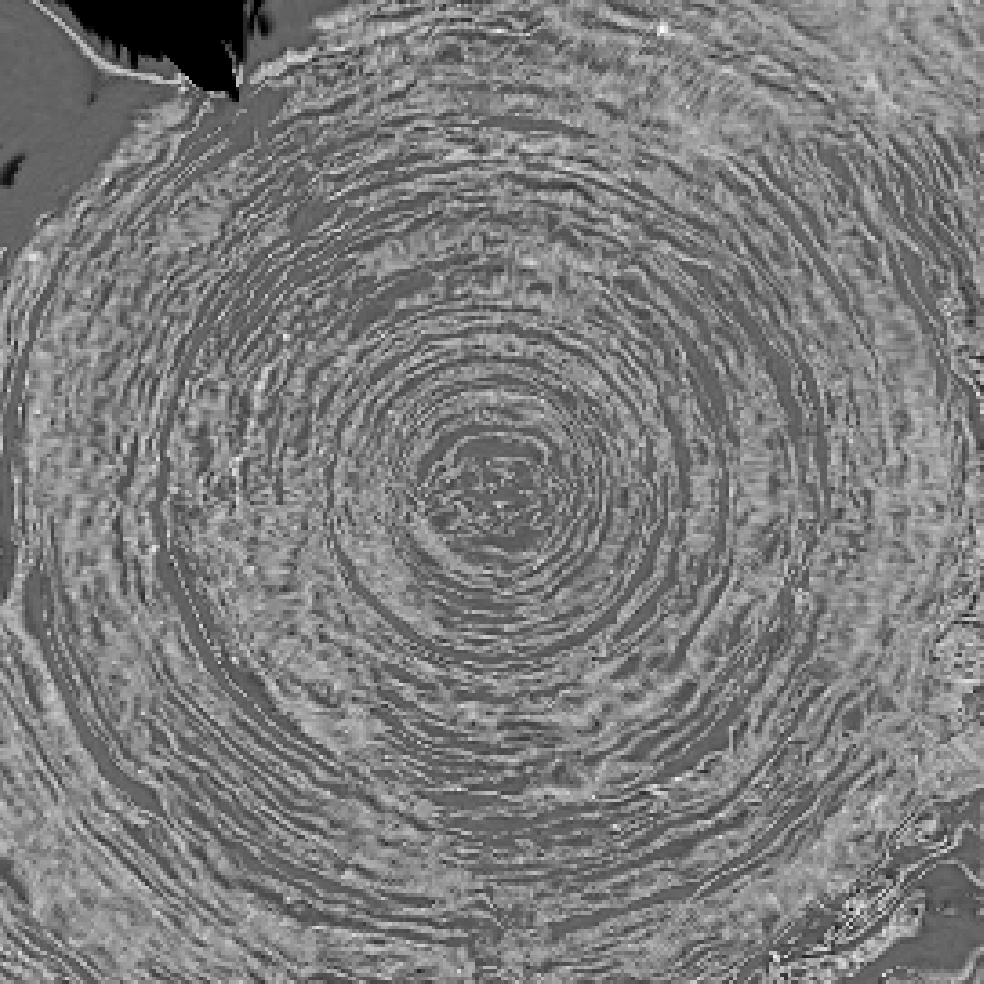}
    
    \vspace{2pt}
    \caption{}
    \label{fig:s1-undeformed-slices}
\end{subfigure}%
    
    \vspace{-10pt}
    \caption{
    \textbf{(a)}
    Cross-sections of \phpf{} with the spiral overlaid, before optimization (top) when the diffeomorphism is an identity transform; and after optimization (bottom). During fitting the spiral adapts to follow the distorted scroll surface.
    \textbf{(b)}
    Undeformed scroll cross-sections, given by transforming the scan with the inverse diffeomorphism.
    The distorted windings in (a) become near-circular in (b).
    }
    
\end{figure*}

\begin{figure*}
    \centering
    \includegraphics[width=0.33\linewidth]{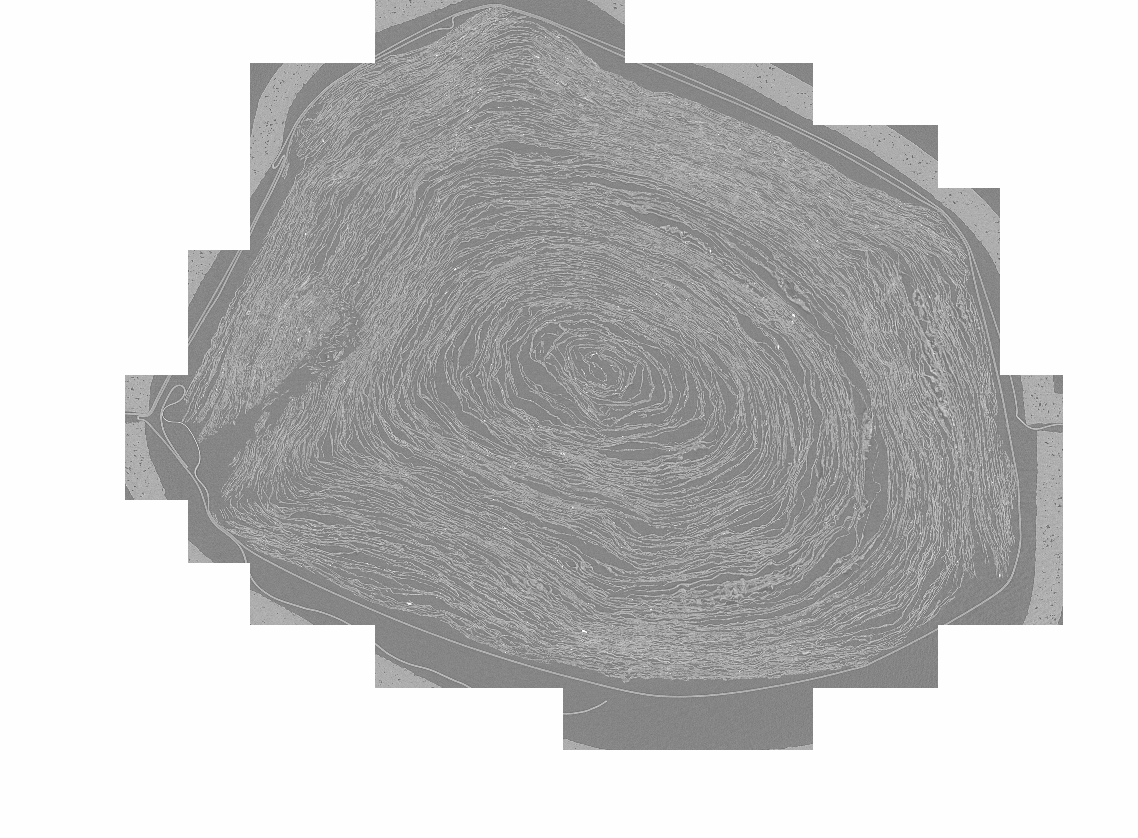}\includegraphics[width=0.33\linewidth]{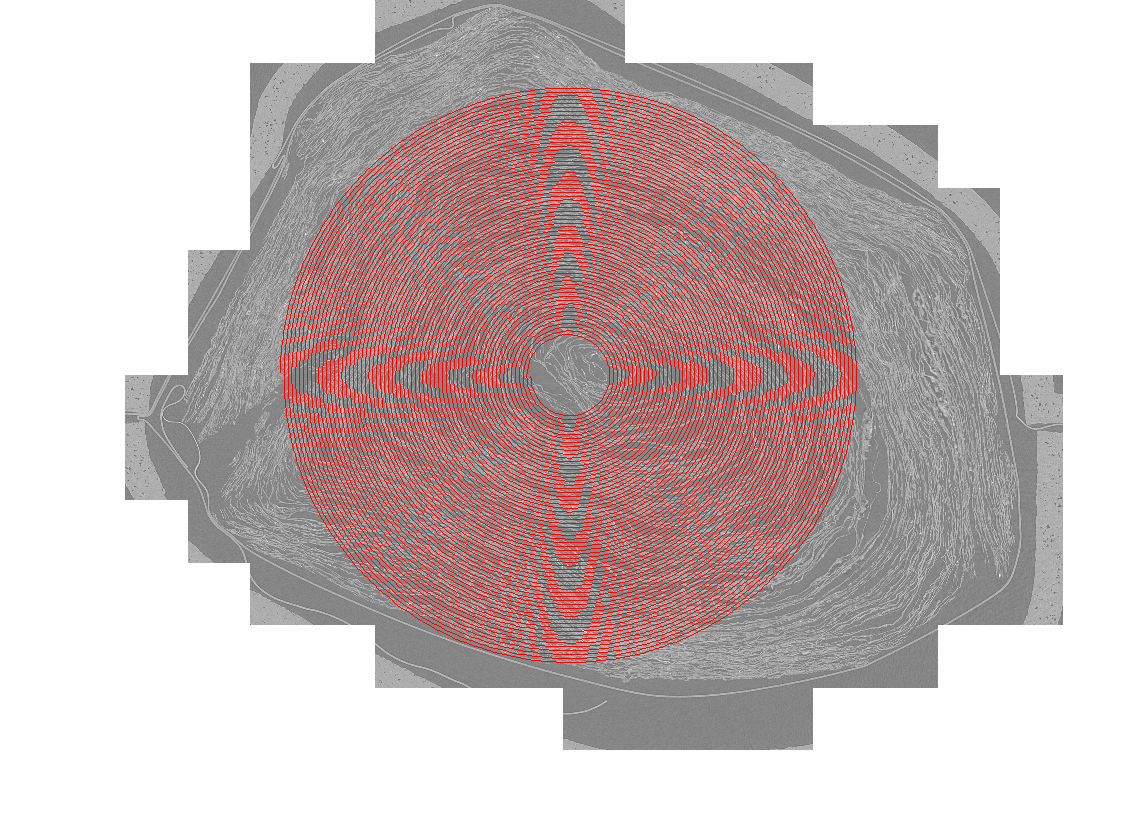}\includegraphics[width=0.33\linewidth]{figures/s5-spiral-sections/video_s1050_01.png}
    
    \includegraphics[width=0.33\linewidth]{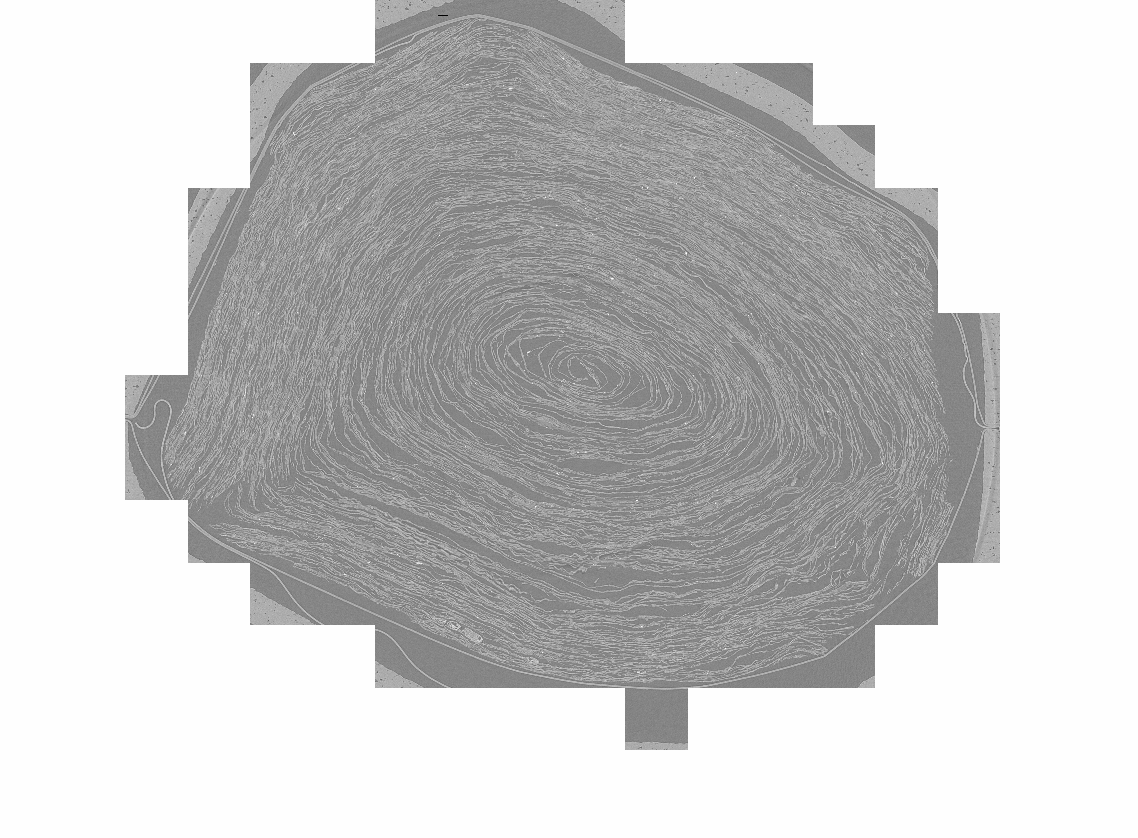}\includegraphics[width=0.33\linewidth]{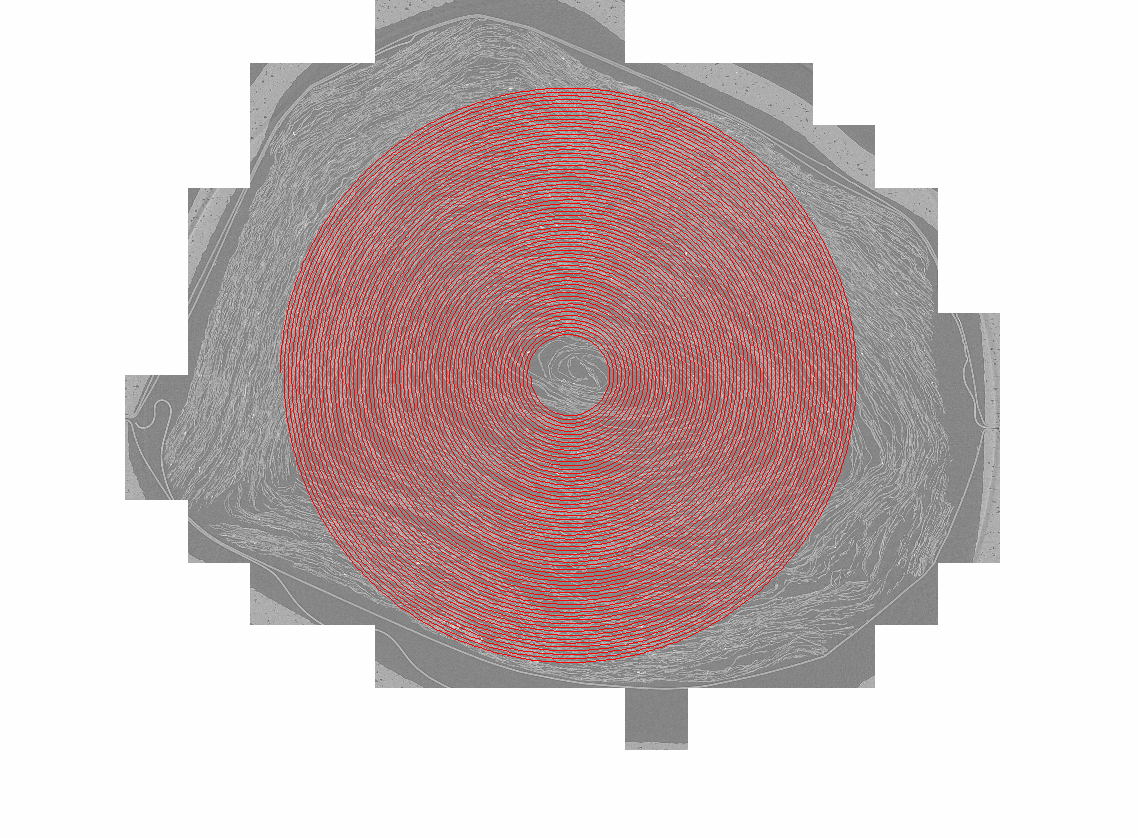}\includegraphics[width=0.33\linewidth]{figures/s5-spiral-sections/video_s1500_01.png}
    
    \includegraphics[width=0.33\linewidth]{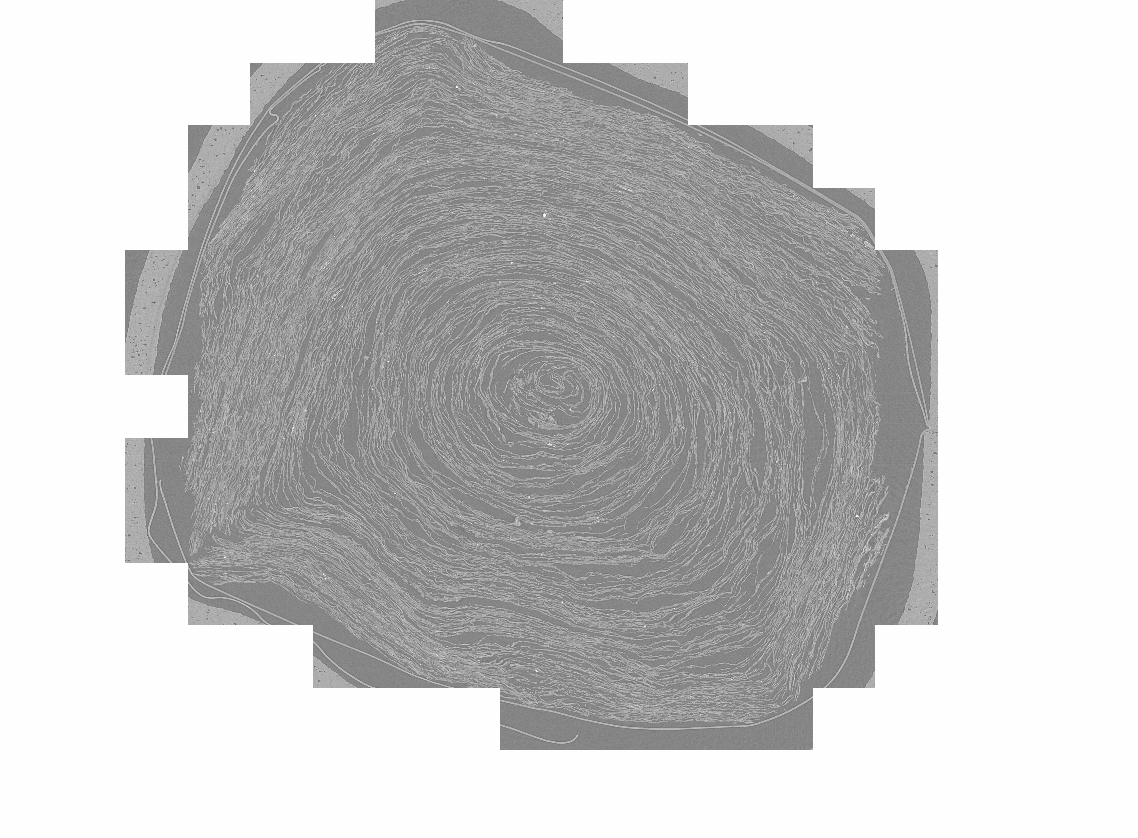}\includegraphics[width=0.33\linewidth]{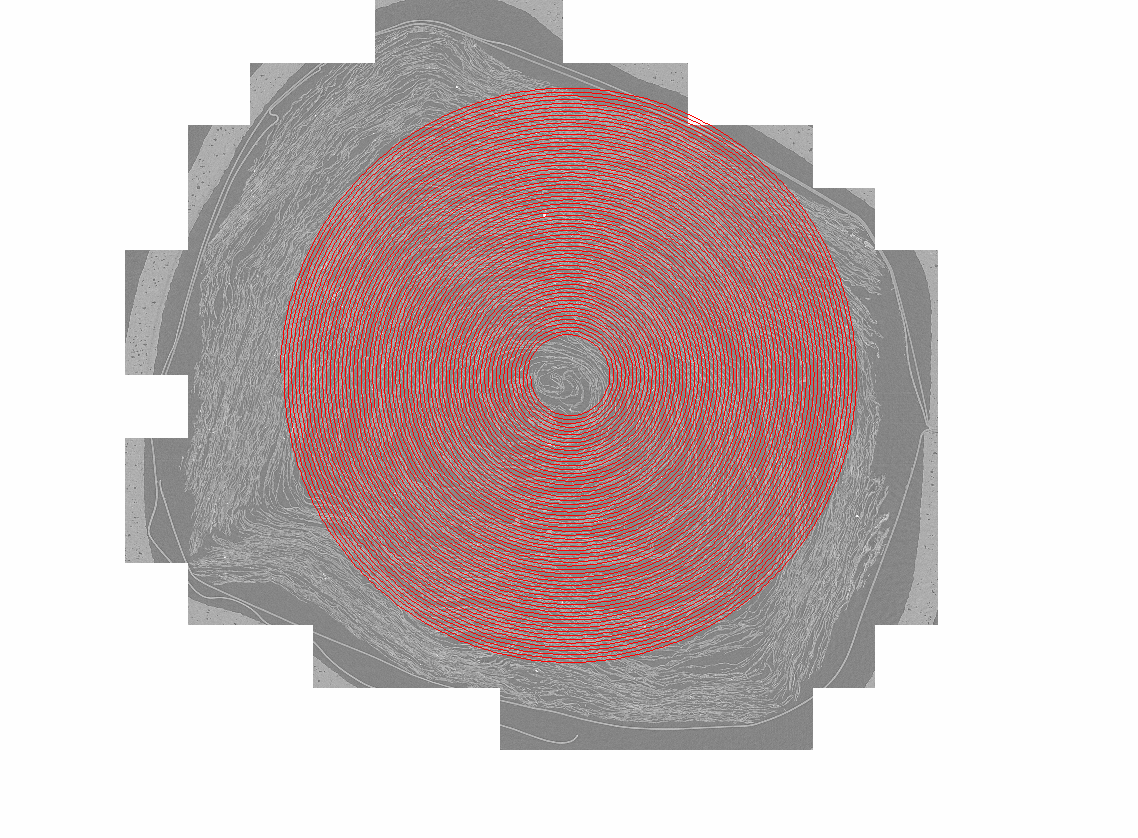}\includegraphics[width=0.33\linewidth]{figures/s5-spiral-sections/video_s2450_01.png}

    \caption{
    Cross-sections of \phost{}.
    \textbf{Left:} original slices, showing the highly deformed, tightly packed windings.
    \textbf{Middle:} spiral overlaid before optimization, when the diffeomorphism is an identity transform and the winding rate $\omega$ is arbitrarily chosen.
    \textbf{Right:} spiral overlaid after optimization. During fitting the spiral adapts to follow the distorted scroll surface.
    }
    \label{fig:s5-spiral-initial-final}
\end{figure*}

\begin{figure*}
    \centering
    \includegraphics[width=0.33\linewidth]{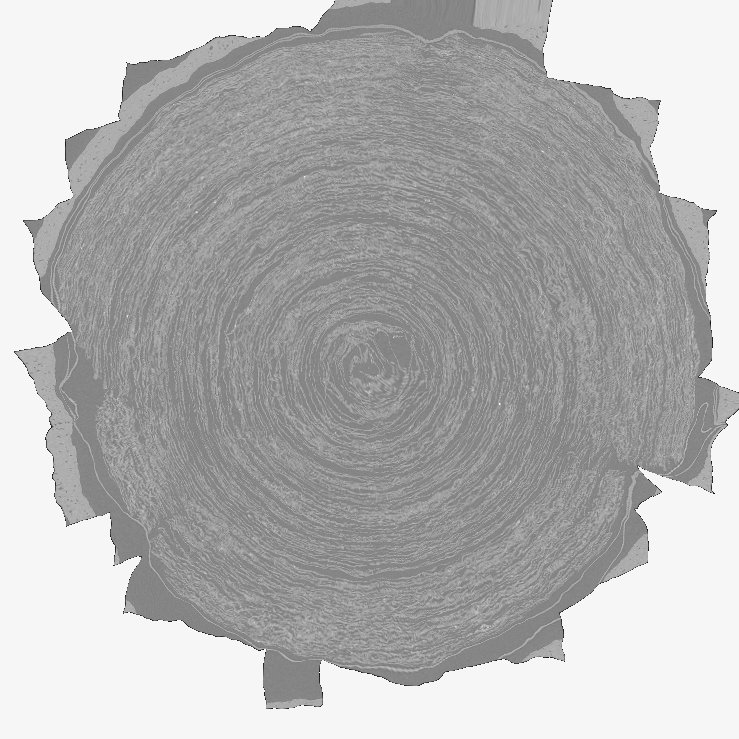}\includegraphics[width=0.33\linewidth]{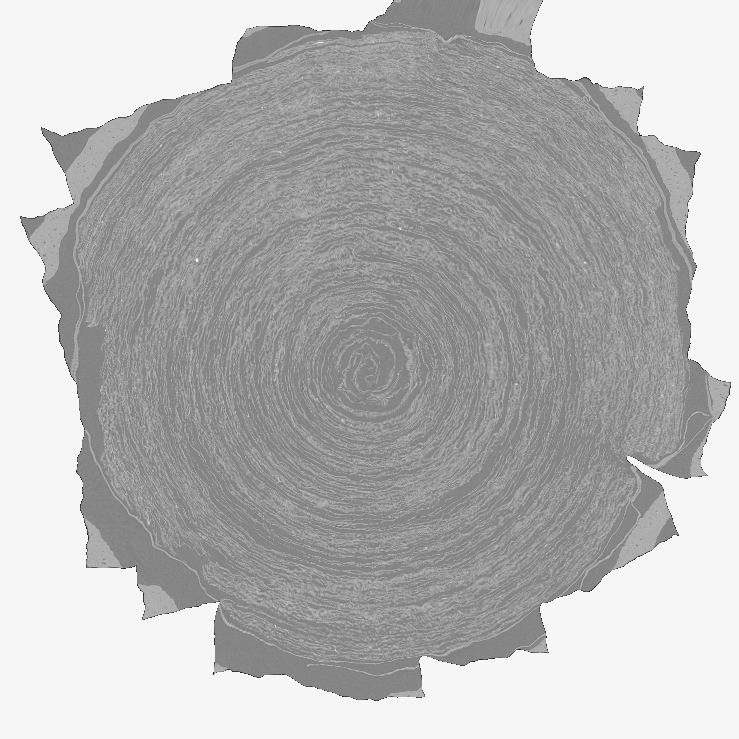}\includegraphics[width=0.33\linewidth]{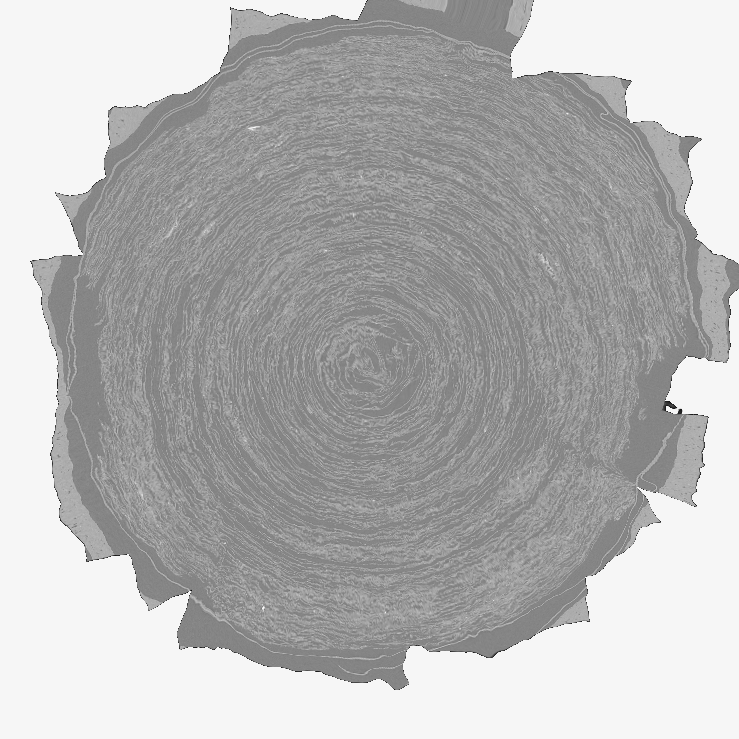}
    
    \caption{
    Undeformed cross-sections from \phost{}, given by transforming the scan with the inverse diffeomorphism.
    The distorted windings in Figure~\ref{fig:s5-spiral-initial-final} become near-circular.
    }
    \label{fig:s5-undeformed}
\end{figure*}

Figure~\ref{fig:s1-spiral-initial-final} shows cross-sections of \phpf{}, with the original undeformed spiral (before optimization) overlaid (top), and the fitted version (bottom).
We also display (Figure~\ref{fig:s1-undeformed-slices}) two slices that are transformed by the inverse diffeomorphism to map them back into canonical space, resulting in near-circular concentric windings.
Similarly, Figure~\ref{fig:s5-spiral-initial-final} shows cross-sections of \phost{} before and after fitting, while Figure~\ref{fig:s5-undeformed} shows undeformed slices with circularized windings.

\begin{figure*}
    \includegraphics[width=\textwidth]{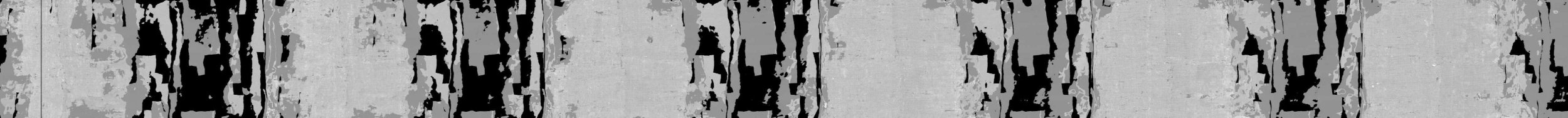}\vspace{2pt}
    \includegraphics[width=\textwidth]{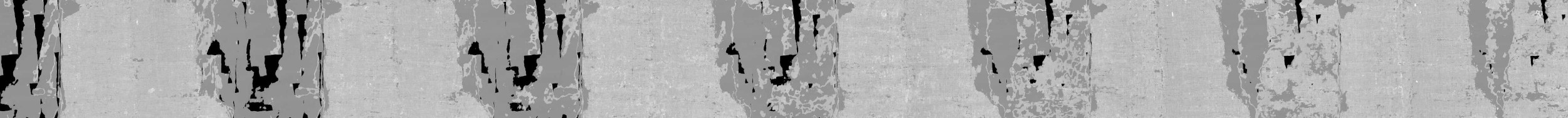}\vspace{2pt}
    \includegraphics[width=\textwidth]{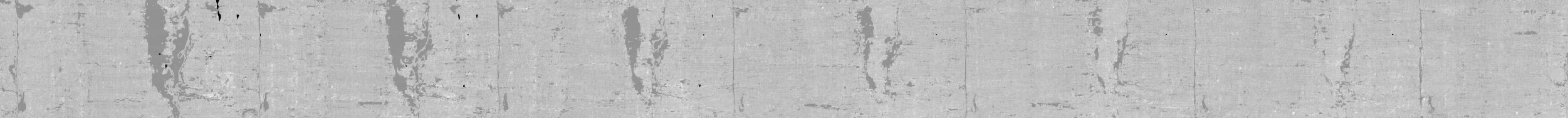}\vspace{2pt}
    \includegraphics[width=\textwidth]{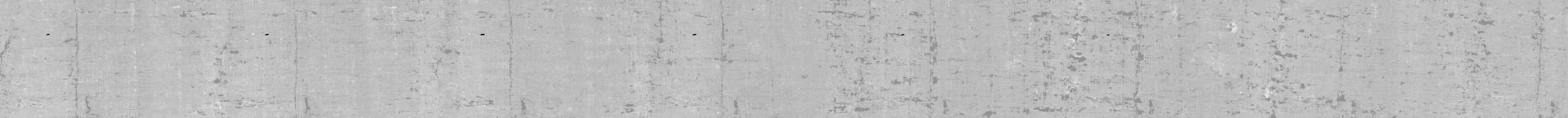}\vspace{2pt}
    \includegraphics[width=\textwidth]{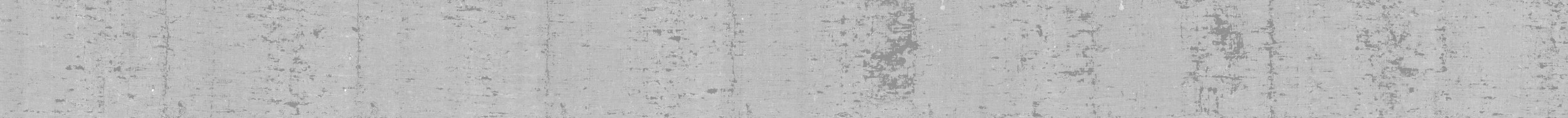}\vspace{2pt}
    \includegraphics[width=\textwidth]{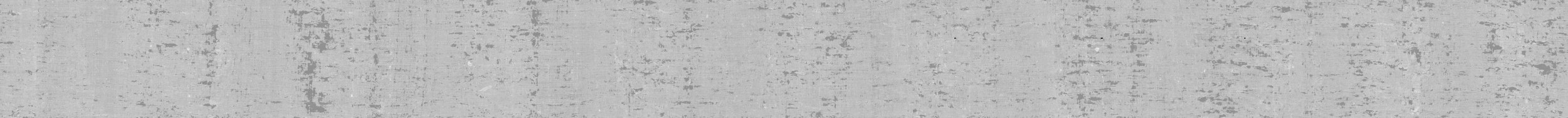}\vspace{2pt}
    \includegraphics[width=\textwidth]{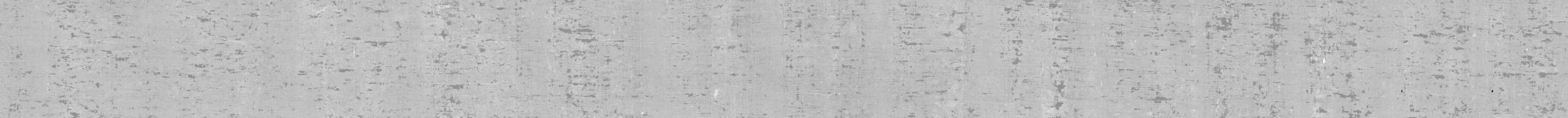}\vspace{2pt}
    \includegraphics[width=\textwidth]{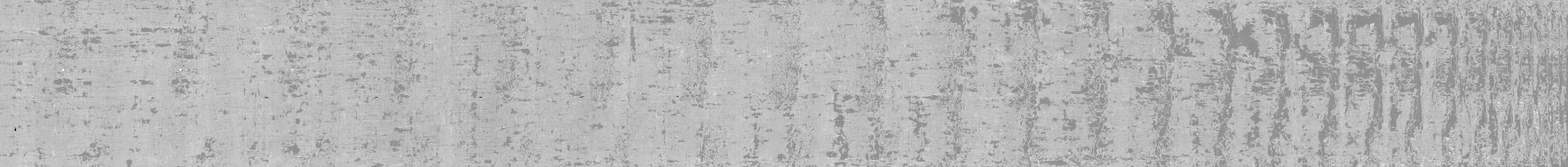}
    \caption{\phost{} unrolled using our method. Note the very long, continuous manifold output. Figures~\ref{fig:s5-ink-1}--\ref{fig:s5-ink-7} show higher-contrast zooms on part of this surface, with ink visible. Black and dark gray parts in the top two rows are where our method has tracked surfaces through regions where the scroll is missing (either burnt away, or damaged by parts flaking off).}
    \label{fig:s5-full-render}
\end{figure*}

In Figure~\ref{fig:s5-full-render} we show the full unrolling of \phost{} produced by our method.
In Figures~\ref{fig:s5-ink-1}--\ref{fig:s5-ink-7} we zoom in on parts of the unwrapped surface, with enhanced contrast.
In each of these views, ink is visible forming Greek characters (light against the darker background) and sometimes complete words.
Note also the individual fibers of papyrus visible forming continuous lines through the view, indicating the our method is correctly following the true sheet surfaces.

\notarxiv{
Alongside this PDF, we also include:
\begin{itemize}
    \item a video interpolating between the idealized scroll with circular cross-section (placed on the centerline of the deformed scroll), and the deformed scroll shape defined by the diffeomorphism
    \item a video showing four cross sectional slices, interpolating between the spiral (red) in its undeformed and deformed states
    \item a high-resolution version of the virtually unrolled scroll, with the texture of the papyrus visible
\end{itemize}
}

\newcommand{\sfink}[2]{
\begin{figure*}
    \includegraphics[width=\linewidth]{figures/s5-renders-and-ink/#1.jpg}
    \caption{Zoom on our virtual unrolling PHerc.172, with ink clearly visible as bright areas forming Greek characters}
    \label{#2}
\end{figure*}
}
\sfink{dg20-composite_c038}{fig:s5-ink-1}
\sfink{dg2-composite_c040}{fig:s5-ink-2}
\sfink{dg2-composite_c044}{fig:s5-ink-3}
\sfink{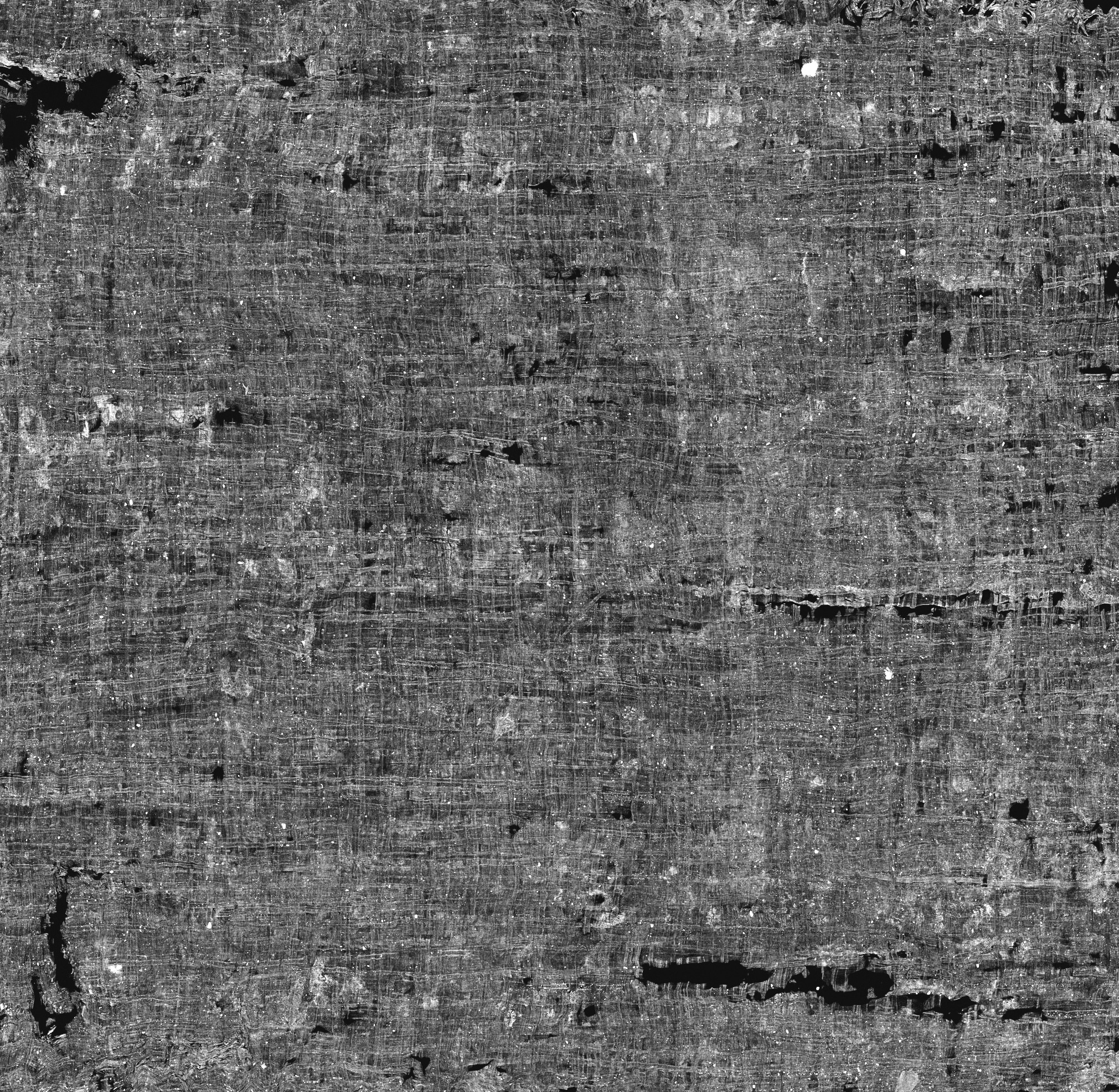}{fig:s5-ink-4}
\begin{figure*}
\centering%
    \includegraphics[width=0.9\linewidth]{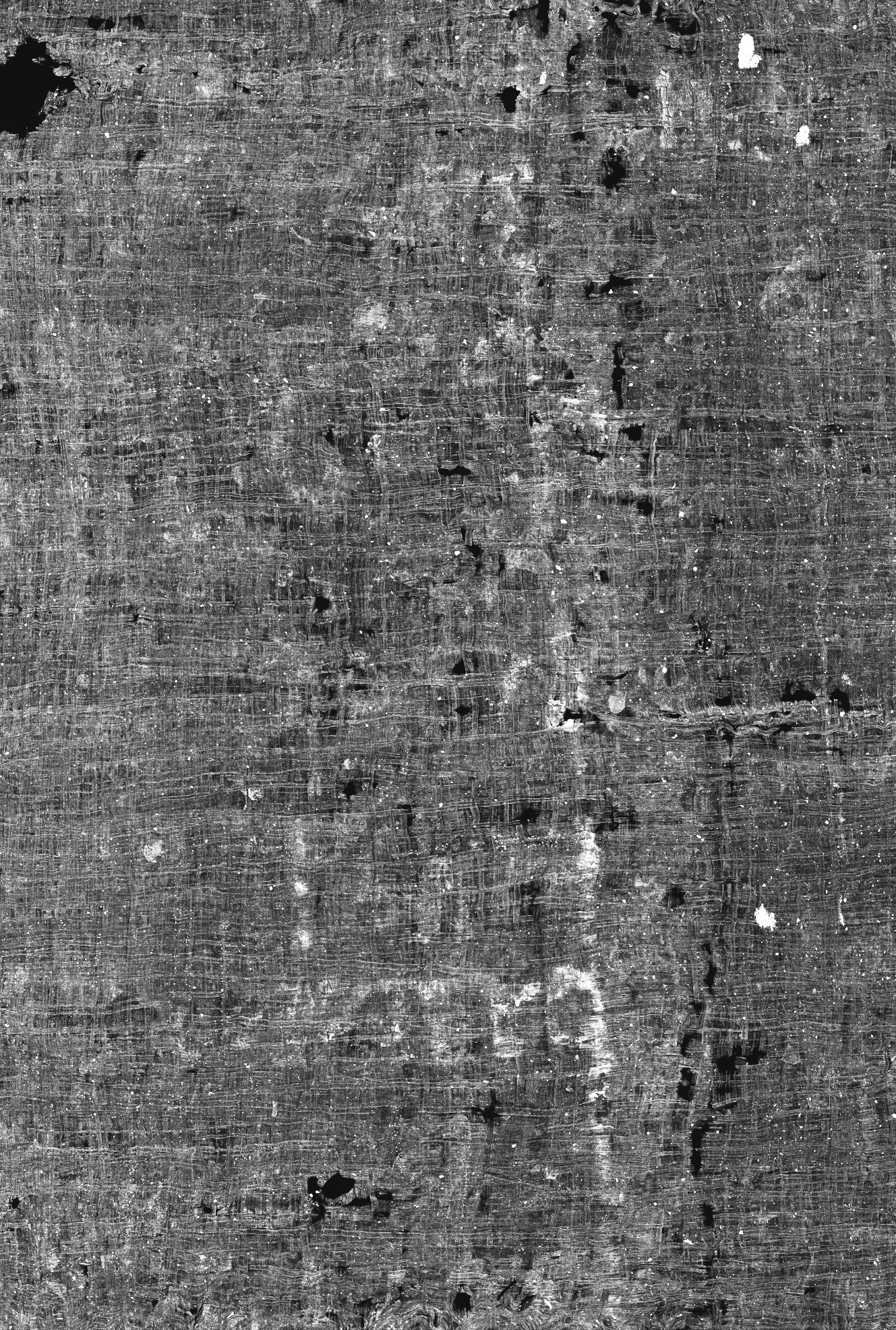}
    \caption{Zoom on our virtual unrolling PHerc.172, with ink clearly visible as bright areas forming Greek characters}
    \label{fig:s5-ink-5}
\end{figure*}
\sfink{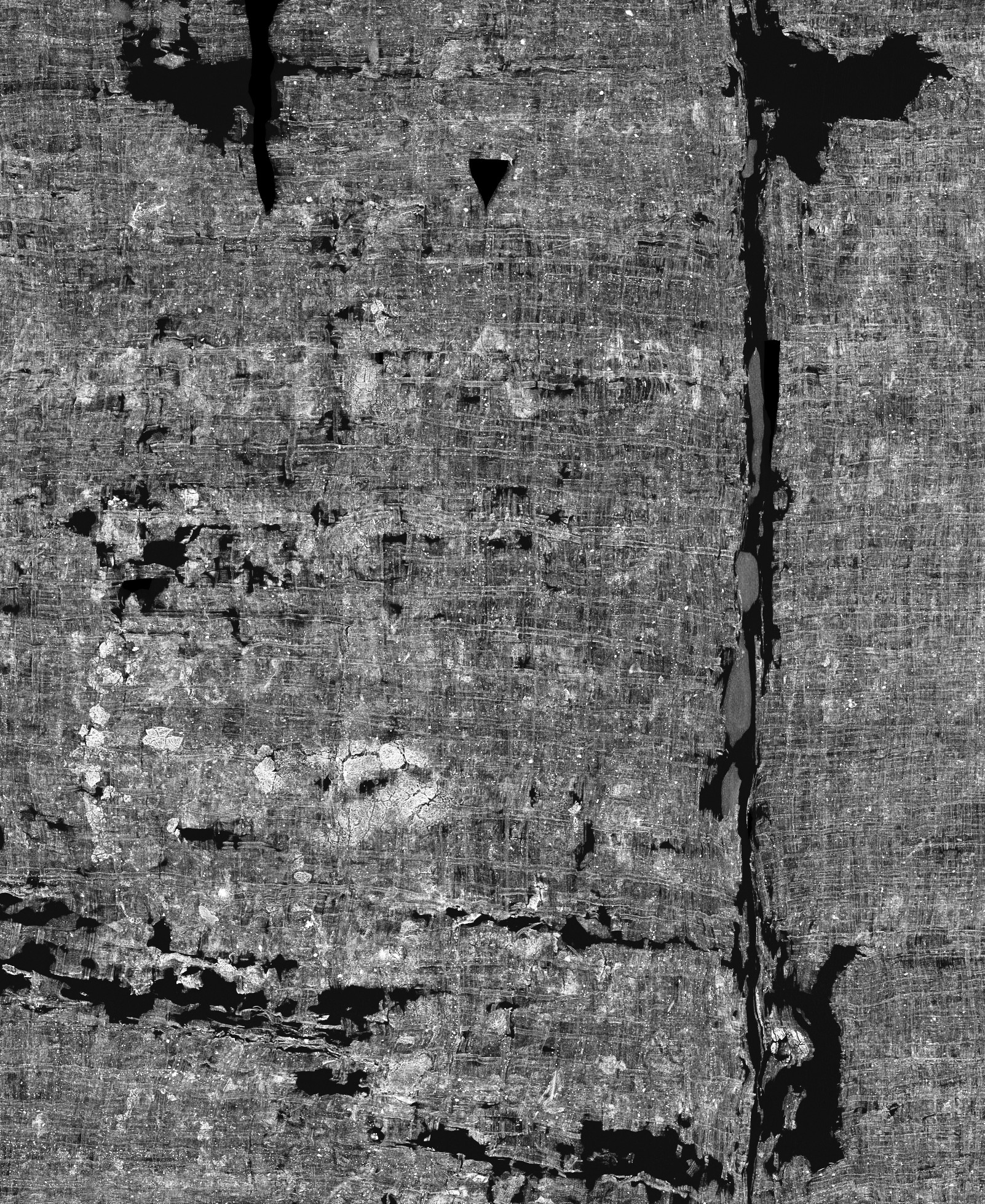}{fig:s5-ink-6}
\sfink{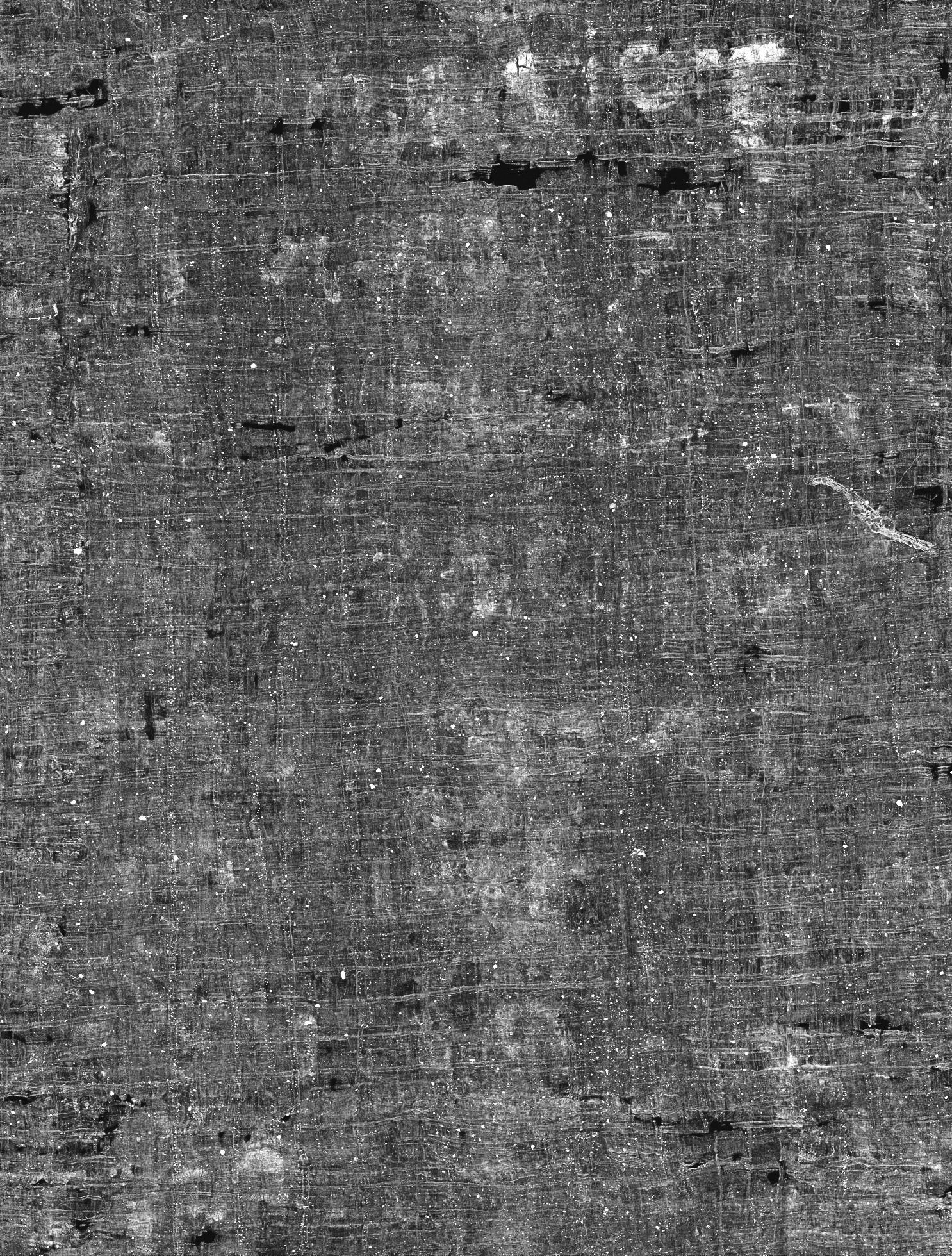}{fig:s5-ink-7}

\end{document}